\documentclass{article}

\PassOptionsToPackage{numbers, compress}{natbib}

\usepackage[preprint]{neurips_2025}

\usepackage[utf8]{inputenc} 
\usepackage[T1]{fontenc}    
\usepackage{hyperref}       
\usepackage{url}            
\usepackage{booktabs}       
\usepackage{amsfonts}       
\usepackage{nicefrac}       
\usepackage{microtype}      
\usepackage[dvipsnames,table]{xcolor}         

\usepackage[pdftex]{graphicx}
\usepackage{amsmath}
\usepackage{mathtools}
\usepackage{amssymb}
\usepackage{multirow}
\usepackage{multicol}
\usepackage{makecell}
\usepackage{subcaption}
\usepackage{enumitem}

\def\ie{\emph{i.e.}}
\def\eg{\emph{e.g.}}
\def\etc{\emph{etc}}

\newcommand{\name}[0]{ZeroGUI}

\usepackage{listings}
\usepackage{tcolorbox}
\tcbuselibrary{listingsutf8}
\usepackage{xcolor}
\usepackage{tabularx}

\definecolor{lightgray}{gray}{0.95}
\tcbuselibrary{breakable}
\lstdefinestyle{promptstyle}{
  backgroundcolor=\color{lightgray},
  basicstyle=\ttfamily\footnotesize,
  breaklines=true,
  frame=single,
  columns=fullflexible,
  showstringspaces=false
}

\usepackage{authblk}

\usepackage[misc]{ifsym}

\title{ZeroGUI: Automating Online GUI Learning \\ at Zero Human Cost}

\author[ ]{Chenyu Yang$^{2,1}$\thanks{Equal contribution. $^{\dagger}$ Project lead. This work is done when Chenyu Yang, Shiqian Su, Shi Liu, Xuan Dong, Yue Yu, Xuehui Wang and Hao Li are interns at Shanghai Artificial Intelligence Laboratory. \Letter\ Corresponding to Weijie Su <suweijie@pjlab.org.cn> and Xizhou Zhu <zhuxizhou@tsinghua.edu.cn>.}}
\author[2,1*]{Shiqian Su}
\author[1*]{Shi Liu}
\author[2,1*]{Xuan Dong}
\author[2,1*]{Yue Yu}
\author[1*,${\dagger}$\Letter]{Weijie Su}
\author[3,1]{\\Xuehui Wang}
\author[4]{Zhaoyang Liu}
\author[]{Jinguo Zhu}
\author[1]{Hao Li}
\author[5,1]{Wenhai Wang}
\author[1]{Yu Qiao}
\author[2,1 \Letter]{\\Xizhou Zhu}
\author[2,1]{Jifeng Dai}
\affil[1]{Shanghai Artificial Intelligence Laboratory \quad $^2$Tsinghua University}
\affil[3]{Shanghai Jiao Tong University \quad $^4$Hong Kong University of Science and Technology}
\affil[5]{The Chinese University of Hong Kong}
\affil[ ]{\tt\small \{yangcy23,ssq24,x-dong21,yuyue21\}@mails.tsinghua.edu.cn,}
\affil[ ]{\tt\small \{liushi,suweijie,zhujinguo,qiaoyu\}@pjlab.org.cn, \{zhuxizhou,daijifeng\}@tsinghua.edu.cn, \quad wangxuehui@sjtu.edu.cn,}
\affil[ ]{\tt\small \{zyliumy,lihaothu\}@gmail.com,whwang@ie.cuhk.edu.hk,}

\begin{document}

\maketitle


\begin{abstract}
The rapid advancement of large Vision-Language Models (VLMs) has propelled the development of pure-vision-based GUI Agents, capable of perceiving and operating Graphical User Interfaces (GUI) to autonomously fulfill user instructions. However, existing approaches usually adopt an offline learning framework, which faces two core limitations: (1) heavy reliance on high-quality manual annotations for element grounding and action supervision, and (2) limited adaptability to dynamic and interactive environments. To address these limitations, we propose \textbf{\name{}}, a scalable, online learning framework for automating \textbf{GUI} Agent training at \textbf{Zero} human cost. Specifically, \name{} integrates (i) VLM-based automatic task generation to produce diverse training goals from the current environment state, (ii) VLM-based automatic reward estimation to assess task success without hand-crafted evaluation functions, and (iii) two-stage online reinforcement learning to continuously interact with and learn from GUI environments. Experiments on two advanced GUI Agents (UI-TARS and Aguvis) demonstrate that \name{} significantly boosts performance across OSWorld and AndroidLab environments.
The code is available at \url{https://github.com/OpenGVLab/ZeroGUI}.

\end{abstract}

\vspace{-0.5em}
\section{Introduction}
\label{sec:intro}
\vspace{-0.5em}

GUI agents are designed to perceive and interact with Graphical User Interfaces (GUIs).
Early methods achieved this by building pipelines or relying on structured inputs such as HTML or DOM trees.
Recently, the emergence of large Vision-Language Models (VLMs)~\cite{wang2024cogvlm, Qwen2-VL, Qwen2.5-VL} has enabled the development of end-to-end, pure-vison-based agents\cite{hong2024cogagent, cheng2024seeclick, xu2024aguvis, qin2025ui}, capable of perceiving GUI screenshots and performing actions such as clicking, scrolling, or typing to complete user-provided task instructions. These agents have demonstrated strong potential across various applications, including digital task automation, intelligent copilots, human-computer interaction, \etc. 

\begin{figure}[t]
    \centering
    \vspace{-1em}
    \includegraphics[width=1.0\linewidth]{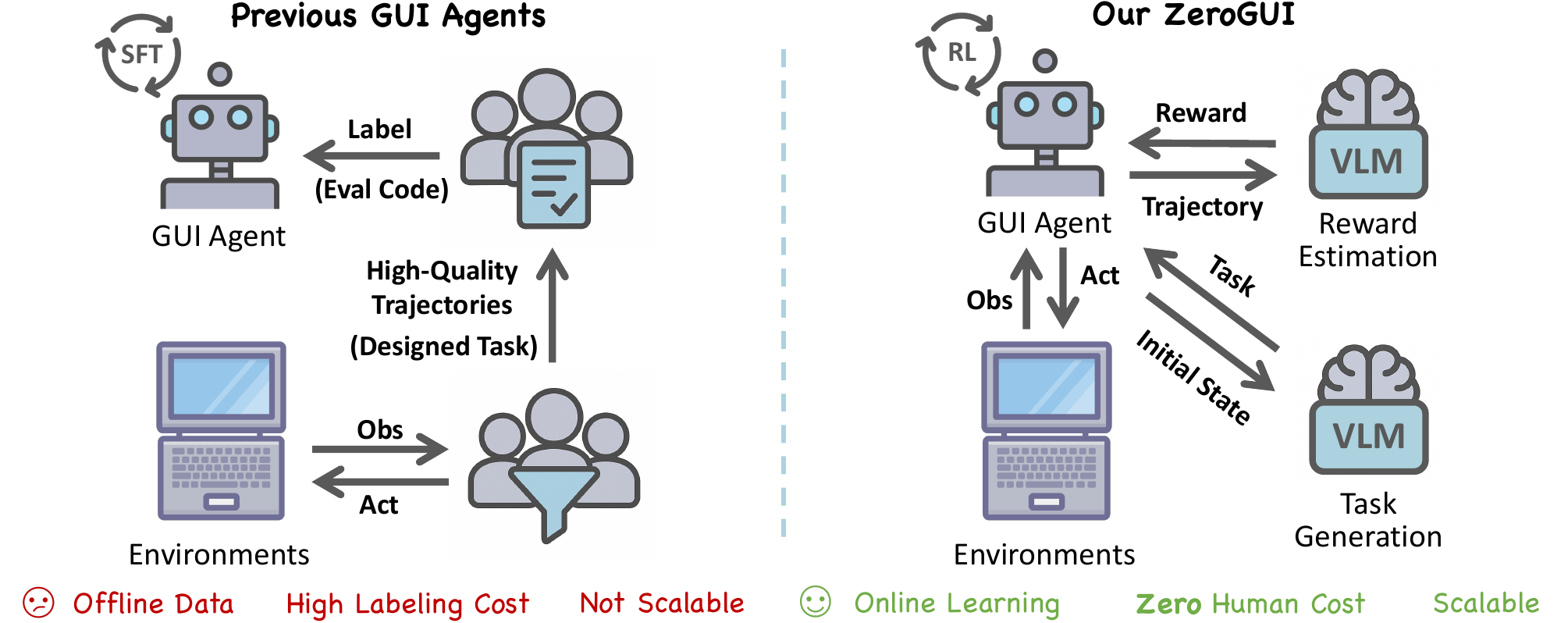}
    \caption{\textbf{Left: Existing Offline Training Framework for GUI Agents} incurs high human costs, relying on manually collected and annotated interaction trajectories, typically under a supervised fine-tuning (SFT) paradigm.
    \textbf{Right: Our \name{}} is a scalable online learning framework with automated task generation and reward estimation at zero human cost. A VLM proposes diverse tasks, which are executed by the agent; the agent then receives VLM-based rewards and updates its policy via reinforcement learning (RL).}
    \label{fig:overview}
    \vspace{-1.5em}
\end{figure}

Despite these progress, as shown in Fig.~\ref{fig:overview}, existing approaches usually adopt an offline learning framework, which presents two fundamental limitations:
\emph{(1) they heavily rely on high-quality human annotations for both GUI grounding~\cite{cheng2024seeclick, wu2024atlas, li2025autogui} (i.e., identifying screen elements) and action trajectories~\cite{chen2024guicourse,lu2024gui,xu2024aguvis,qin2025ui} (i.e., a sequence of actions to complete a task).}
These human-annotated labels are expensive, time-consuming, and difficult to scale across diverse platforms and tasks.
\emph{(2) they fall short in adapting to dynamic and interactive environments.}
GUIs in the real world are non-stationary and uncertain: elements may shift, disappear, or behave differently depending on the system's state. 
Existing agents often overfit to static or narrowly defined tasks and struggle to generalize in open-ended scenarios.

To overcome these limitations, online learning, where GUI agents are continuously updated through interaction with GUI environments, is a desirable approach but remains challenging.
Most existing environments, such as OSWorld~\cite{xie2024osworld} and AndroidLab~\cite{xu2024androidlab}, only provide a test set consisting of manually crafted tasks and verification functions. 
Constructing a training set with diverse tasks and associated success verifiers by hand is expensive and not scalable.
Furthermore, in real-world environments, novel scenarios or tasks often lack ground-truth annotations, making it difficult to provide direct supervisory signals for agent learning.

To develop a scalable online learning framework, we focus on automating the construction of tasks and their corresponding success verifiers in GUI environments.
This is feasible with the help of advanced VLMs, because they have been trained with large-scale GUI-related data and excel at understanding GUI elements, actions, and state transitions.
They can assess task completion and propose relevant tasks based on observed information.
In addition, when training GUI agents, it is sufficient to evaluate the encountered states rather than exhaustively covering all possible scenarios, which significantly reduces the complexity of automated task verification.

Specifically, we propose \name{}, a fully automated online training framework in which GUI agents continuously interact with GUI environments to complete automatically generated tasks and update their policies using annotation-free rewards.
As illustrated in Fig.~\ref{fig:overview}, \name{} consists of the following components:
\emph{(1) VLM-Based Automatic Task Generation}, which proposes a large and diverse set of training tasks based on random initial states.
\emph{(2) VLM-Based Automatic Reward Estimation}, which predicts binary rewards that indicate task success, serving as supervision signals. The estimator leverages the trajectories of GUI agents as input, eliminating the need for hand-crafted task verifiers.
\emph{(3) Two-stage Online Reinforcement Learning}, involves a training stage on generated tasks followed by a test-time adaptation stage.
We adapt the RL framework to support multi-step interactions between the GUI agent and the environment.

We apply our \name{} to two advanced VLM-based GUI agents (UI-TARS~\cite{qin2025ui} and Aguvis~\cite{xu2024aguvis}) and leverage both desktop (OSWorld~\cite{xie2024osworld}) and mobile (AndroidLab~\cite{xu2024androidlab}) environments for evaluation.
Experiments show that our \name{} yields significant improvements in task success rates.
Training on generated tasks extends the agent's capability coverage, while test-time training helps the agent adapt to test tasks.
In particular, on OSWorld, \name{}-UI-TARS-7B achieves 14\% relative improvements and \name{}-Aguvis-7B achieves 63\% relative improvements.

In summary, our contributions are as follows:
\begin{itemize}[leftmargin=1.5em]
    \vspace{-0.7em}
    \item We propose \name{}, a fully automated online learning framework that enables GUI agents to improve through interaction with GUI environments, eliminating the need for collecting and labeling offline training data.
    \vspace{-0.2em}
    \item We design automatic VLM-based task generation and reward estimation, which generate training tasks and provide supervisory rewards in GUI environments without human annotations.
    \vspace{-0.2em}
    \item We introduce a two-stage reinforcement learning strategy.
    In the first stage, training on generated tasks builds the agent’s general capabilities. In the second stage, test-time training enables the agent to adapt to target test tasks.
    \vspace{-0.2em}
    \item The proposed \name{} significantly improves task success rates across multiple GUI environments and generalizes well to different base models.
\end{itemize}

\section{Related Work}

\textbf{GUI Agents} are AI systems aimed to perceive, understand, and act upon graphical user interfaces. Early systems \cite{yao2023react, wen2024autodroid} heavily relied on structured representations such as HTML, DOM trees, but recent progress in Vison Language Models (VLMs) has enabled a shift toward purely vision-based approaches. 
However, due to the small size and visual variability of UI elements, general-purpose VLMs still struggle with accurate grounding.
To mitigate this, several works \cite{zheng2024gpt, gou2024navigating, wan2024omniparser, yu2025omniparser} incorporate specialized UI parsers to assist proprietary VLMs \cite{achiam2023gpt, hurst2024gpt, anthropic2024claude35sonnet, google2024gemini15pro}. Other efforts \cite{hong2024cogagent, cheng2024seeclick, wu2024atlas, lin2024showui, xu2024aguvis, yang2024aria, li2025autogui, qin2025ui} focus on building large-scale grounding datasets via manual labeling or automated pipelines for supervised fine-tuning.

Beyond grounding, the long-horizon and high-variability nature of GUI tasks makes planning another key challenge. To this end, works such as \cite{chen2024guicourse, lu2024gui} collect expert trajectories across platforms for imitation learning. 
Aguvis\cite{xu2024aguvis} enhances these datasets with VLM generated chain-of-thought annotations, while UI-TARS\cite{qin2025ui} introduces both positive and negative samples to facilitate self-reflection and error correction via direct preference optimization.
Given the high cost of collecting high-quality demonstrations, recent methods \cite{lu2025ui, xia2025gui, liu2025infigui} explore reinforcement fine-tuning with rule-based rewards and limited expert data. 
Nevertheless, several studies \cite{bai2024digirl, qi2024webrl} report that models trained solely on static trajectories often struggle to generalize in dynamic, real-world environments. To improve adaptability, works like \cite{qi2024webrl, liu2024autoglm} train outcome reward models (ORMs) in dynamic and interactive environments \cite{zhou2023webarena, liu2024visualagentbench} to support more robust RL-based adaptation. 
However, many interactive GUI environments \cite{xie2024osworld, rawles2024androidworld} lack curated training task sets, making ORM training difficult. To address this, we propose \textit{VLM-based Automatic Task Generation} and \textit{VLM-based Automatic Reward Estimation} methods, enabling scalable RL training in dynamic and interactive environments and laying the foundation for a more generalizable GUI agent training framework.

\textbf{Interactive Environments for GUI Agents.} Environments for GUI agents can be broadly categorized into non-interactive and interactive settings. Non-interactive environments (e.g., \cite{deng2023mind2web,zheng2024gpt,liu2024visualwebbench} for the web domain; \cite{li2020mapping,sun2022meta,hsiao2022screenqa,rawles2023androidinthewild,zhang2024android,li2024effects,lu2024gui,chai2024amex} for Android; and cross-platform settings like \cite{kapoor2024omniact,chen2024guicourse}) are typically static and predefined. Interactive environments (e.g., \cite{gao2024assistgui,bonatti2024windows,zhao2025worldgui} for Linux or Windows; \cite{shi2017world,liu2018reinforcement,yao2022webshop,zhou2023webarena,koh2024visualwebarena,he2024webvoyager,lu2024weblinx,drouin2024workarena,pan2024webcanvas} for web; \cite{wang2024mobile,rawles2024androidworld,xu2024androidlab} for Android; and cross-platform frameworks such as \cite{xie2024osworld,xu2024crab,liu2024visualagentbench}) facilitate autonomous perception and action, enabling agents to operate in more realistic and dynamic settings.

Interactive environments for GUI agents typically consist of an operating system platform (e.g., desktop OS or Android emulator), an action execution interface, and an observation module. Agents interact with the environment through a predefined action space, or using PyAutoGUI on desktops and Android Debug Bridge (ADB) on mobile platforms. The environment provides feedback in the form of screenshots and UI structure data, enabling agents to perceive both visual and structured information for decision-making. Evaluation metrics in this dynamic environments are primarily based on manually predefined rules or hardcoded scripts, which are complex and lack scalability.

\textbf{RL for Post-Training.} Reinforcement learning methods such as PPO\cite{schulman2017proximal} and DPO\cite{rafailov2023direct} have been widely used in the post-training of Large Language Models (LLMs), particularly within the RLHF framework\cite{ouyang2022training}. Recently, RL methods such as GRPO\cite{shao2024deepseekmath}, DAPO\cite{yu2025dapo}, and Dr.GRPO\cite{liu2025understanding} have demonstrated strong effectiveness in improving reasoning abilities of LLMs, as exemplified by works like \cite{guo2025deepseek, team2025kimi}. Building on these developments, a series of studies \cite{yang2025r1,zhou2025r1,huang2025vision,deng2025boosting,peng2025lmm,liu2025visual,shen2025vlm, chen2025r1v} extend RL approaches to the post-training of VLMs, mainly focusing on mathematical reasoning and other general vision tasks where annotated labels are easily accessible.

In parallel, emerging paradigms such as test-time reinforcement learning\cite{zuo2025ttrl} attract increasing attention, aiming to enable effective RL training on unlabeled data. Our work further advances this line of research by applying test-time RL to VLM post-training in dynamic, interactive GUI environments. Specifically, we adapt the GRPO algorithm to multi-step settings and enhance training stability through tuning the KL regularization term.

\section{\name{}}

Existing offline methods rely on carefully collected trajectories and designed tasks, limiting their scalability and adaptability. To enable online adaptation with vision-language models, we introduce \name{}, an automatic online training framework at zero human cost.

\noindent\textbf{Formulation.} The completion of a GUI task can be formulated as a Markov Decision Process (MDP)~\cite{bellman1957markovian}, denoted by $(\mathcal{S}, \mathcal{A}, \mathcal{R}, \mathcal{T})$.
Given a task instruction $I$, the GUI agent interacts with the environment. At each step $t$, the agent predicts an action $a_t$ according to its policy:
\begin{equation}
    a_t \sim \pi_\theta\left(a_t|I,s_t\right), \ \ \ \ s_t=(o_t, h_t)
    \label{eq:agent}
\end{equation}
where the agent's state $s_t$ combines the current observation $o_t$ and history information $h_t$ containing previous observations and actions.
The process terminates when encountering a terminate action or reaching maximum number of steps, resulting in a trajectory:
\begin{equation}
    \tau = \{I, (o_1, a_1), (o_2, a_2), \dots, (o_T, a_T)\}
    \label{eq:traj}
\end{equation}

\begin{figure}
    \centering
    \vspace{-1em}
    \includegraphics[width=1.0\linewidth]{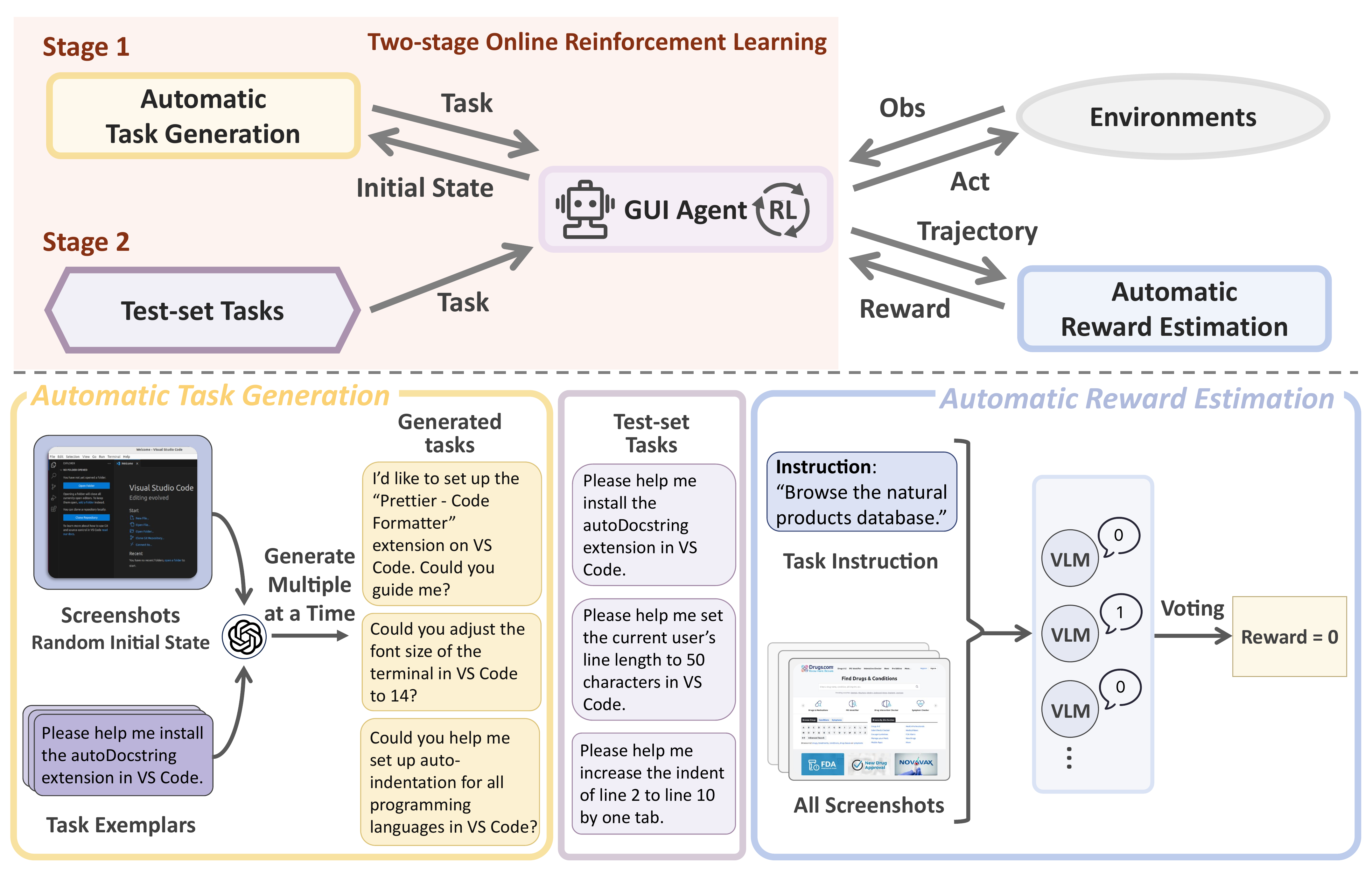}
    \caption{\textbf{Top: Overview of \name{}.} It adopts a \textbf{Two-stage Online Reinforcement Learning} paradigm. In the first stage, tasks are automatically generated by a VLM, while in the second stage, tasks are drawn from the test set. These tasks are executed by the GUI agent. After each interaction, a reward is assigned automatically by the VLM based on the agent's trajectory, and the policy network is updated via reinforcement learning. \textbf{Bottom left: Automatic Task Generation.} The VLM receives a random initial screenshot and a set of task exemplars to generate diverse novel tasks. \textbf{Bottom right: Automatic Reward Estimation.} The final reward is obtained via majority voting of multiple VLM evaluations based on all screenshots of the trajectory.}
    \label{fig:framework}
    \vspace{-1em}
\end{figure}

\noindent\textbf{Framework Overview.} As illustrated in Fig.~\ref{fig:framework}, \name{} consists of three key components: automatic task generation, automatic reward estimation, and a two-stage online reinforcement learning process.
In the first stage, a VLM automatically generates a set of training tasks, and the agent is trained using rewards estimated by a VLM-based evaluator.
In the second stage, the agent performs test-time training on the test set without ground-truth labels, receiving rewards solely from the same VLM-based evaluator.
The following sections describe each component of our \name{} in detail.

\subsection{Automatic Task Generation}

Current offline training methods heavily rely on high-quality human annotations for GUI grounding and trajectory data with manually designed task instructions. To enable online training without human supervision, it is crucial to develop a scalable and automated task generation pipeline.

A key challenge in task generation is to ensure generalization, particularly given the limited number of evaluation samples in existing GUI agent benchmarks. Our generated tasks must not only align with the operational constraints of the target environments but also exhibit sufficient diversity to cover a broad behavioral space. To address this, we propose the following prompting designs:

\emph{(1) Example-Guided Prompting.} Prompt with a combination of instruction exemplars and randomly sampled initial state screenshots, which guide the model toward environment-specific and realistic task proposals. 

\emph{(2) Multi-Candidate Generation.} In each generation step, we request multiple task candidates simultaneously, encouraging the model to produce a diverse set rather than overfitting to a narrow task style. 

Fig.~\ref{fig:framework} illustrates this generation process and showcases representative generated tasks with test tasks from OSWorld as reference.

To further train the agents to recognize unachievable goals and provide appropriate feedback, we prompt the VLM to generate a subset of infeasible tasks.
Such tasks are intentionally unsolvable within the environment and require the agent to explicitly output ``FAIL'' response.

\subsection{Automatic Reward Estimation}

Existing interactive environments typically use script-based verifiers running within them to determine task success (\eg checking file contents or system states). 
These verifiers often involve complex commands and logic to cover all possible cases, which heavily relies on manual implementation and debugging.
Thus, setting up such verifiers for large-scale generated training tasks is both costly and unnecessary.

To support general task verification or reward estimation in online training, we employ a vision-language model (VLM) to assign binary rewards to trajectories. However, VLM-based assessment is often imperfect, \eg, it may overlook details or suffer from hallucinations, leading to incorrect labeling.
Among the two error types, \ie, false positives and false negatives, our experiments show that false positives have a greater impact (Sec.~\ref{sec:ablation}). 
This is likely because positive rewards offer more informative signals for policy improvement, and too many false positives can distort the improvement. In contrast, negative rewards often provide less targeted guidance, making the effect of false negatives relatively minor.

Therefore, as illustrated in Fig.~\ref{fig:framework}, the reward estimator focuses on reducing false positives and improving precision with the following designs:

\emph{(1) All screenshots in the trajectory are included.}
The success of some tasks can only be determined by changes in the environment before and after an action, so all screenshots are needed.

\emph{(2) The agent’s responses are \textbf{excluded}}. They may contain hallucinations of success, even when the task actually fails. Such content can mislead the VLM to give false positive rewards.

\emph{(3) A voting mechanism is adopted.} 
The VLM is queried multiple times, and the reward is assigned based on either majority agreement or a stricter unanimous agreement (i.e., $R=1$ only if all outputs indicate success).
This voting strategy further reduces the risk of false positives.

\subsection{Two-stage Online Reinforcement Learning}
\label{subsec:two_stage_online_rl}

With both automatic task generation and reward estimation mechanisms in place, the GUI agent can perform online learning by continuously interacting with the GUI environment and updating its policy guided by the rewards.
Furthermore, since our reward estimator does not rely on internal environment states or ground-truth labels, it can also provide rewards for test tasks, enabling test-time adaptation.
To this end, we introduce a two-stage training strategy:

\emph{(1) Training on generated tasks.} The agent learns fundamental capabilities from generated tasks.\\
\emph{(2) Test-time training.} The agent adapts to target test tasks with rewards from the reward estimator.

Reinforcement learning (RL) is adopted for this two-stage online training.
We start from the Group Relative Policy Optimization (GRPO)~\cite{shao2024deepseekmath}, which eliminates the need for an additional value function and is effective for the post-training of LLMs and VLMs in other scenarios~\cite{guo2025deepseek, yang2025r1, chen2025r1v}.
To adapt the original GRPO algorithm to the online RL of GUI agents, we propose the following modifications:

\emph{(1) Extend the optimization objective to multi-step trajectories.}
For a given task $I$, we follow GRPO to sample a group of trajectories $\{\tau^{(i)}\}_{i=1}^G$ and obtain their rewards $\{R^{(i)}\}_{i=1}^G$.
The advantage $\hat{A}^{(i)}$of the trajectory $\tau^{(i)}$ is computed by normalizing the rewards within the group (Eq.~\ref{eq:loss_2}).
The difference comes that in original GRPO, each sample is a single generated sequence, while in our setting, each trajectory consists of multiple action prediction sequences from model-environment interaction (Eq.~\ref{eq:traj}).
For a trajectory $\tau^{(i)}$, we assign each prediction sequence $a_t^{(i)}$ with advantage $\hat{A}^{(i)}$ and compute its objective as follow:
\begin{equation}
\label{eq:loss}
\mathcal{J}^{(i)}_t(\theta)=\frac{1}{|a_t^{(i)}|}\sum_{k=1}^{|a_t^{(i)}|}\left(\min \left(r^{(i)}_{t,k}(\theta)\hat{A}^{(i)}, \text{clip}\left(r^{(i)}_{t,k}(\theta), 1-\epsilon, 1+\epsilon\right)\hat{A}^{(i)}\right)-\beta D_{\text{KL}}(\pi_\theta||\pi_{\text{ref}})\right)
\end{equation}
where
\begin{equation}
\label{eq:loss_2}
r^{(i)}_{t,k}(\theta)=\frac{\pi_\theta\left(a_{t,k}^{(i)}\big|I, s^{(i)}_t, a_{t,<k}^{(i)}\right)}{\pi_{\theta_{\text{old}}}\left(a_{t,k}^{(i)}\big|I, s^{(i)}_t, a_{t,<k}^{(i)}\right)},\ \ \ \ 
\hat{A}^{(i)}=\frac{R^{(i)}-\text{mean}\left(\{R^{(i)}\}_{i=1}^G\right)}{\text{std}\left(\{R^{(i)}\}_{i=1}^G\right)}
\end{equation}
where $t$ denotes the $t$-th step in the trajectory. The action $a_t^{(i)}$ is sampled according to Eq.~\ref{eq:agent}, and $k$ indicates the $k$-th output token. The final objective is the average over all sequences:
\begin{equation}
    \mathcal{J}(\theta)=\frac{1}{\sum_{i=1}^G T^{(i)}}\sum_{i=1}^G\sum_{t=1}^{T^{(i)}}\mathcal{J}^{(i)}_t(\theta)
\end{equation}

\emph{(2) Modify the KL loss term for better training stability.} 
GRPO uses the k3-estimator~\cite{schulman2020kl} for the KL loss, \ie, $D_{\text{KL}}^{\text{GRPO}}=\pi_{\text{ref}}/\pi_\theta-\log \pi_{\text{ref}}/\pi_\theta-1$. 
However, we find that it can cause large gradients and is prone to overflow or underflow.
We replace it with the k2-estimator~\cite{schulman2020kl}, \ie, per-token MSE loss, which provides more stable gradients and avoids numerical overflow:
\begin{equation}
   D_{\text{KL}}(\pi_\theta||\pi_{\text{ref}}) = \frac{1}{2}\left(\log\pi_\theta\left(a_{t,k}^{(i)}\big|I, s^{(i)}_t, a_{t,<k}^{(i)}\right) - \log\pi_{\text{ref}}\left(a_{t,k}^{(i)}\big|I, s^{(i)}_t, a_{t,<k}^{(i)}\right)\right)^2
\end{equation}
Experiments in Sec.~\ref{sec:ablation} confirm that this modification improves training stability. 
Besides, removing the KL constraint may cause policy distribution drift, so simply dropping this loss term is not a desirable option.
Further derivations and analyses are provided in the appendix.

\vspace{-0.5em}
\section{Experiment}
\vspace{-0.3em}

\subsection{Experiment Settings}
\label{sec:exp_setting}

\subsubsection{Evaluation Environments and Metrics}

\noindent\textbf{OSWorld.} OSWorld~\cite{xie2024osworld} is a benchmark built upon computer environment designed for evaluating multi-modal agents on complex real-world tasks. 
It comprises 369 tasks that span web applications, desktop software, and OS-level operations. 
Among them, 30 tasks (8.1\% of the test set) are infeasible by design to assess the ability to detect deprecated or hallucinated features.
We report evaluation results on both the full test set and the subset of feasible tasks (\ie, excluding the infeasible ones).

Our evaluation is conducted on the Ubuntu platform from OSWorld with screenshot-only
mode.
The screen size is $1920\times1080$ and the maximum number of steps is limited to 15.
To reduce the influence of network instability and environmental variability, we report the mean and standard deviation of scores over 4 runs.
Additionally, we incorporate the following metrics for further analysis:
(1) $\text{pass}@k$: the expected proportion of tasks the model can solve within $k$ trials, reflecting its potential capacity coverage. (2) $\text{all-pass}@k$: the expected proportion of tasks the model completes in all $k$ trials, indicating the consistency in performance.
The unbiased estimators~\cite{chen2021evaluating} are given by: 
\begin{equation}
\text{pass}@k \coloneqq \mathbb{E}_{x_i\sim D}\left[1-\frac{\binom{n-c_i}{k}}{\binom{n}{k}}\right], \ \ \ \ 
\text{all-pass}@k \coloneqq \mathbb{E}_{x_i\sim D}\left[\frac{\binom{c_i}{k}}{\binom{n}{k}}\right]
\end{equation}
\vspace{-0.5em}

where each task $x_i$ is tested $n$ times and $c_i$ is the number of
correct samples. We set $n=8$, $k=4$, and the sampling temperature as 0.5 to estimate these two metrics.

\noindent\textbf{AndroidLab.}
AndroidLab~\cite{xu2024androidlab} is an interactive Android environment that includes the Android system and 9 offline-deployable apps (e.g., Clock, Calendar).
It comprises 138 test tasks, which are categorized into two types: operational tasks and query-detecting tasks.
Operational tasks involve completing goals through operations and are evaluated by predefined rules.
Query-detecting tasks require the model to extract information and return a text answer, scored by GPT.
We observe that the GPT-based evaluation for certain tasks is not fully reliable, so we report evaluation results on both the full test set and the subset of operation tasks.
While existing methods have been evaluated in XML or SoM modes, we implement a screenshot-only setting to support our model and test the corresponding baseline.
Success rate (SR) and sub-goal success rate (Sub-SR) are caculated as the metrics.
\vspace{-0.3em}
\subsubsection{Implementation Details}
\vspace{-0.3em}
For task generation, we use GPT-4o~\cite{hurst2024gpt} to generate
10 tasks for OSWorld and 5 tasks for AndroidLab at the same time. 
In total, more than 4,000 Ubuntu-based tasks and 225 Android-based tasks are generated. For training, 725 Ubuntu tasks and 175 Android tasks are randomly sampled from the generated pool, which is approximately twice the size of their respective test sets.

For reward estimation, Qwen2.5-VL-32B~\cite{Qwen2.5-VL} is deployed locally for efficiency.
We query the VLM 4 times with a temperature of 1.0 and use unanimous agreement voting to determine the reward.

For training, we choose UI-TARS-7B-DPO~\cite{qin2025ui} and Aguvis-7B~\cite{xu2024aguvis} as our base models.
We use the AdamW~\cite{loshchilov2017decoupled} optimizer with a constant learning rate of 2e-6.
For GRPO~\cite{shao2024deepseekmath}, we set the group size $G=64$ and the KL coefficient $\beta=0.1$.
We adopt DAPO~\cite{yu2025dapo} dynamic sampling, which filters out tasks with accuracy equal to 1 or 0.
For each rollout step, sampling continues until 16k sequences are collected, followed by a single gradient update.
We train 1 epoch for both generated tasks and test-time tasks.
Ablation studies are conducted on the \emph{Daily} domain of OSWorld (including three apps: Chrome, Thunderbird, and VLC Player) to reduce experimental burden.
More details are provided in the appendix.

\setlength{\tabcolsep}{3pt}
\setlength{\doublerulesep}{2\arrayrulewidth}
\renewcommand{\arraystretch}{1.1}
\begin{table}[t]
    \centering
    \small
    \caption{\textbf{Test results on OSWorld benchmark.} Test settings and metrics are described in Sec.~\ref{sec:exp_setting}. The success rates are reported in ``mean$\pm$std''. Absolute and relative improvements with respect to the base model are highlighted in \textcolor{ForestGreen}{green}. * reported results taken from previous papers.}
    \label{table:main_osworld}
    \resizebox{1.0\linewidth}{!}{
    \begin{tabular}{l|ccc|ccc}
        \toprule 
        \multirow{2}{*}{Model} & \multicolumn{3}{c|}{Test set} & \multicolumn{3}{c}{Feasible subset} \\
        & SR & $\text{pass}@4$ & $\text{all-p}@4$ & SR & $\text{pass}@4$ & $\text{all-p}@4$ \\
        \midrule
        GPT-4o~\cite{hurst2024gpt} & 5.0* & - & - & - & - & - \\
        Gemini-Pro-1.5~\cite{google2024gemini15pro} & 5.4* & - & - & - & - & - \\
        Claude Computer-Use~\cite{anthropic2024claude35sonnet} & 14.9* & - & - & - & - & - \\
        OpenAI Operator~\cite{openai_operator_2025} & 19.7* & - & - & - & - & - \\
        \midrule
        CogAgent-9B-20241220~\cite{hong2024cogagent} & 8.1* & - & - & - & - & - \\
        Aguvis-72B~\cite{xu2024aguvis} & 10.3* & - & - & - & - & - \\
        UI-TARS-72B-DPO~\cite{qin2025ui} & 22.7* & - & - & - & - & - \\
        \midrule
        Aguvis-7B~\cite{xu2024aguvis} & 3.0$\pm$0.4 & 7.2 & 1.4 & 2.4$\pm$0.5 & 6.5 & 0.6 \\
        \rowcolor{Gray!15}
        + \name{} (Gen. task only) & 4.1$\pm$0.3 \scriptsize\textcolor{ForestGreen}{(+1.1, 37\%)} & 8.0 & 1.3 & 3.6$\pm$0.4 \scriptsize\textcolor{ForestGreen}{(+1.2, 50\%)} & 7.6 & 0.7 \\
        \rowcolor{Gray!30}
        \textbf{+ \name{}} & 4.9$\pm$0.4 \scriptsize\textcolor{ForestGreen}{(+1.9, 63\%)} & 8.3 & 1.8 & 4.5$\pm$0.4 \scriptsize\textcolor{ForestGreen}{(+2.1, 88\%)} & 7.9 & 1.1 \\
        \midrule
        UI-TARS-7B-DPO~\cite{qin2025ui} & 17.7$\pm$1.1 (18.7*) & 25.5 & 9.4 & 11.3$\pm$0.6 & 18.5 & 4.9 \\
        \rowcolor{Gray!15}
        + \name{} (Test-time only) & 18.2$\pm$0.9 \scriptsize\textcolor{ForestGreen}{(+0.5, 3\%)} & 26.4 & 8.6 & 14.4$\pm$0.8 \scriptsize\textcolor{ForestGreen}{(+3.1, 27\%)} & 21.7 & 6.9 \\
        \rowcolor{Gray!15}
        + \name{} (Gen. task only) & 18.2$\pm$1.3 \scriptsize\textcolor{ForestGreen}{(+0.5, 3\%)} & 27.8 & 7.5 & 14.7$\pm$1.0 \scriptsize\textcolor{ForestGreen}{(+3.4, 30\%)} & 22.1 & 5.6 \\
        \rowcolor{Gray!30}
        \textbf{+ \name{}} & 20.2$\pm$1.0 \scriptsize\textcolor{ForestGreen}{(+2.5, 14\%)} & 28.0 & 9.6 & 15.8$\pm$0.5 \scriptsize\textcolor{ForestGreen}{(+4.5, 40\%)} & 22.2 & 7.3 \\
        \bottomrule
    \end{tabular}}
\vspace{-1em}
\end{table}

\vspace{-0.5em}
\subsection{Main Results}
\subsubsection{OSWorld}
\vspace{-0.3em}

We evaluate our proposed \name{} on the OSWorld benchmark, and the results are shown in Tab.~\ref{table:main_osworld}. Other existing approaches are also listed as reference.

(1) Compared to the base models, our proposed \name{} leads to significant improvements in task success rate, especially on the feasible subset.
Specifically, for UI-TARS-7B-DPO, we achieve a +2.5, 14\% improvement on all tasks and +4.5, 40\% on the feasible subset.
For Aguvis-7B, although the base model performs poorly, our method still yields gains of +1.9, 63\% and +2.1, 88\%, respectively, with even greater relative improvements.
This demonstrates the effectiveness and generalization of our self-improving online training framework.

(2) Both of our training stages: generated task training and test-time training, contribute to performance gains. 
The $\text{pass}@4$ and $\text{all-pass}@4$ metrics further reveal their complementary roles. 
generated task training improves $\text{pass}@4$ significantly, indicating that large-scale and diverse generated tasks help to extent the model's capability coverage.
Test-time training mainly boosts $\text{all-pass}@4$, suggesting that the behavior consistency of the model is enhanced after being adapted to the target tasks.
Notably, using only test-time training underperforms the two-stage setup, highlighting that generated training provides a beneficial ability foundation that allows RL in the next stage to unlock more tasks and gain more informative rewards.

(3) The improvement on the full test set is smaller than on the feasible subset (\eg, +2.5 vs. +4.5 in the average success rate for UI-TARS-7B-DPO), indicating a decrease in the detection of infeasibility. This may be due to two reasons: (a) the VLM lacks detailed knowledge of specific software, making it hard to judge infeasibility; (b) noisy rewards with false positives may cause the model to become overconfident. 
To mitigate this, we included a portion of generated infeasible tasks in the training set, which has greatly alleviated this problem (see Sec.~\ref{sec:ablation}).

\vspace{-0.3em}
\subsubsection{AndroidLab}
\vspace{-0.3em}

We also evaluate \name{} on the AndroidLab benchmark, with the results shown in Tab.~\ref{table:main_androidlab}.

(1) From the SR perspective, \name{} achieves a +2.8 improvement on the operation subset and a +1.8 improvement on the full test set. This demonstrates that the proposed \name{} generalizes well across different interactive GUI environments.

(2) From the Sub-SR perspective, \name{} achieves a +2.9 improvement on the operation subset. Despite leveraging only the overall task rewards, it still yields performance gains in sub-task metrics.

\setlength{\tabcolsep}{2pt}
\setlength{\doublerulesep}{2\arrayrulewidth}
\renewcommand{\arraystretch}{1.1}
\begin{table}[t]
    \centering
    \small
    \vspace{-1em}
    \caption{\textbf{Test results on AndroidLab benchmark.} Test settings and metrics are described in Sec.~\ref{sec:exp_setting}. The success rates are reported in ``mean$\pm$std''.  Absolute improvements with respect to the base model are highlighted in \textcolor{ForestGreen}{green}. * reported results taken from previous papers. Note: we fix some code errors in the original task verifiers, and some baseline methods get higher scores after the correction.
    }
    \label{table:main_androidlab}
    \begin{tabular}{cl|cc|cc}
        \toprule 
        \multirow{2}{*}{Mode} & \multirow{2}{*}{Model} & \multicolumn{2}{c}{Test set} & \multicolumn{2}{|c}{Operation subset}  \\
        & & SR & Sub-SR & SR & Sub-SR \\
        \midrule
        \multirow{4}{*}{SoM} & Gemini-1.5-Pro~\cite{google2024gemini15pro} & 16.7* & 18.4* & - & - \\
        & Claude-3.5-Sonnet~\cite{anthropic2024claude35sonnet} & 34.8 (29.0*) & 38.5 (32.6*) & 34.5 & 38.8 \\
        & GPT-4o~\cite{hurst2024gpt} & 38.0 (31.2*) & 44.3 (35.0*) & 38.4 & 47.3 \\
        & AutoGLM~\cite{liu2024autoglm} & 36.2* & - & - & -\\
        \midrule
        \multirow{3}{*}{Screenshot} & UI-TARS-7B-DPO~\cite{qin2025ui}  & 45.7$\pm$1.52 & 50.5 & 54.6$\pm$1.72 & 61.5 \\
        & \cellcolor{Gray!15} + \name{} (Gen. task only) & \cellcolor{Gray!15}46.4$\pm$2.05 \scriptsize\textcolor{ForestGreen}{(+0.7)} & \cellcolor{Gray!15} 52.0 \scriptsize\textcolor{ForestGreen}{(+1.5)} & \cellcolor{Gray!15} 55.6$\pm$2.06 \scriptsize\textcolor{ForestGreen}{(+1.0)} & \cellcolor{Gray!15} 63.5 \scriptsize\textcolor{ForestGreen}{(+2.0)} \\
        & \cellcolor{Gray!30} \textbf{+ \name{}} & \cellcolor{Gray!30} 47.5$\pm$2.12 \scriptsize\textcolor{ForestGreen}{(+1.8)} & \cellcolor{Gray!30} 52.6 \scriptsize\textcolor{ForestGreen}{(+2.1)} & \cellcolor{Gray!30} 57.4$\pm$2.30 \scriptsize\textcolor{ForestGreen}{(+2.8)} & \cellcolor{Gray!30} 64.4 \scriptsize\textcolor{ForestGreen}{(+2.9)} \\
        \bottomrule
    \end{tabular}
    \vspace{-1em}
\end{table}

\vspace{-0.5em}
\subsection{Ablation Study}
\label{sec:ablation}
\vspace{-0.3em}

\noindent\textbf{Task Generation.}
Tab.~\ref{table:ablation_generate} ablates our designs of task generation.
(1) Removing examples during task generation or generating only one task at a time leads to a drop in test performance.
We attribute this to two factors: providing task examples helps align the distribution of generated tasks with the target domain, while generating multiple tasks increases diversity, which are crucial for training data.
(2) Excluding infeasible tasks results in a sharp decline on the infeasible subset, indicating that such tasks help the model identify unachievable goals and reduces overconfidence.

\setlength{\tabcolsep}{4pt}
\setlength{\doublerulesep}{2\arrayrulewidth}
\renewcommand{\arraystretch}{1.1}
\begin{table}[t]
\centering
\small
\caption{\textbf{Ablations of key components in task generation and reward estimation.} The models are trained and tested on the \emph{Daily} domain of OSworld.``SR'', ``Feas.'', and ``Infeas.'' denotes the success rates on all test tasks, feasible tasks, and infeasible tasks, respectively.
``w/o multi. cand.'': without multiple candidates. ``All Screen.'': all screenshots. ``With Res.'': with agent's response.}
\subfloat[
Task generation
\label{table:ablation_generate}
]{
\centering
\begin{minipage}{0.415\linewidth}{\begin{center}
\begin{tabular}{l|ccc}
    \toprule 
    Method & SR & Feas. & Infeas.\\
    \midrule
    w/o examples & 22.3 & 20.8 & 33.8 \\
    w/o multi. cand. & 24.1 & 22.0 & 40.7 \\
    w/o infeasible & 24.0 & 24.2 & 22.1 \\
    \rowcolor{Gray!30}
    Ours & 27.2 & 25.4 & 41.3 \\
    \bottomrule
\end{tabular}
\end{center}}\end{minipage}
}
\hfill
\subfloat[
Reward estimation
\label{table:ablation_reward}
]{
\centering
\begin{minipage}{0.56\linewidth}{\begin{center}
\begin{tabular}{ccc|cc|c}
    \toprule 
    All Screen. & With Res. & Voting & Precision & Recall & SR\\
    \midrule
    - & - & - & 47.5 & 40.4 & 23.9 \\
    \checkmark & - & - & 53.7 & 61.7 & 25.6 \\
    \checkmark & \checkmark & - & 44.3 & 74.5 & 24.0 \\
    \rowcolor{Gray!30} 
    \checkmark & - & \checkmark & 61.5 & 51.1 & 27.2 \\
    \bottomrule
\end{tabular}
\end{center}}\end{minipage}
}
\vspace{-1em}
\end{table}

\noindent\textbf{Reward Estimation.} We ablate the key designs in the reward estimator.
First, a set of trajectories (UI-TARS-7B-DPO on generated tasks) are randomly selected and manually labeled the ground-truth rewards. 
We then apply different reward estimation methods to this set and evaluate their precision and recall.
In addition, we train separate models using rewards estimated by each method and compare their success rates on the test tasks.

The results are shown in Tab.~\ref{table:ablation_reward}.
(1) Using only the final screenshot instead of all screenshots results in low precision, recall, and test success rate.
(2) Including the agent’s response during reward estimation yields the highest recall but significantly lowers both precision and test success rate, indicating that the VLM is misled by the response and produces many false positives.
(3) Excluding the agent’s response and applying a voting mechanism increase precision while decreasing recall, and also lead to a notable improvement in test success rate.
This suggests that false positive errors have a more detrimental effect on model training.

\noindent\textbf{RL Training.}
To evaluate the effectiveness of our online RL training, we compare it against two baselines: offline rejection sampling fine-tuning (RFT) and online RFT. Results are reported in Tab.~\ref{table:ablation_rl}.
Offline RFT first collects trajectories for all tasks using the base model and fine-tunes only on positive samples.
Its performance is limited due to a distribution mismatch between the collected trajectories and the updated policy, and it fails to leverage rewards from new tasks discovered after policy updates.
Online RFT performs better but still lags behind online RL. This is mainly because RFT discards all negative samples, while RL enables the model to learn from them and avoid repeating past mistakes.

We evaluate the effect of replacing the k3-KL loss in original GRPO with a k2-KL loss. As shown in Fig.~\ref{fig:kl_train_curve}, k2-KL yields higher and more stable training accuracy. 
Test success rates in Tab.~\ref{table:ablation_rl} further validate the superiority of k2-KL in our setting.

\setlength{\tabcolsep}{5pt}
\setlength{\doublerulesep}{2\arrayrulewidth}
\renewcommand{\arraystretch}{1.1}

\vspace{-0.5em}
\begin{minipage}{\textwidth}
\begin{minipage}[c]{0.47\textwidth}
    \centering
    \small
    \captionof{table}{\textbf{Ablations of online RL training and the modified KL loss.} The models are trained and tested on the \emph{Daily} domain of OSworld.}
    \label{table:ablation_rl}
    \begin{tabular}{l|c}
        \toprule 
        Training Method & SR\\
        \midrule
        Offline RFT & 22.4 \\
        Online RFT & 24.5 \\
        \rowcolor{Gray!30}
        Online RL (Ours) & 27.2 \\
        \midrule
        k3-KL (GRPO) & 26.1 \\
        \rowcolor{Gray!30}
        k2-KL (ours) & 27.2 \\
        \bottomrule
    \end{tabular}
\end{minipage}
\hfill
\begin{minipage}[c]{0.48\textwidth}
    \small
    \centering
    \includegraphics[width=0.8\linewidth]{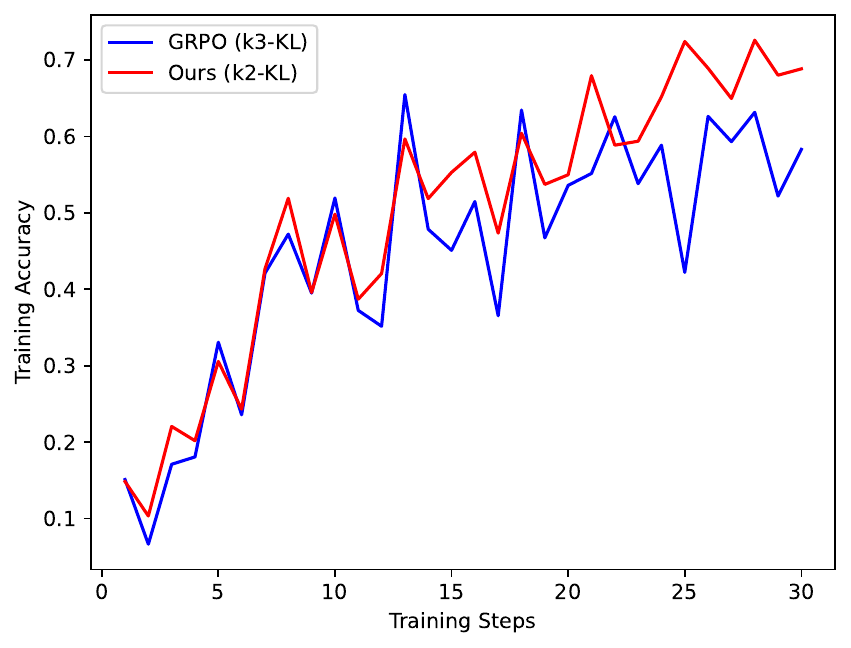}
    \captionof{figure}{Comparison of training accuracies with k3-KL (GRPO) and k2-KL (ours).}
    \label{fig:kl_train_curve}
\end{minipage}
\end{minipage}
\vspace{-0.8em}

\vspace{-0.5em}
\section{Conclusion}
\label{sec:con}
\vspace{-0.5em}

In this work, we present \name{}, a fully automated online learning framework for GUI agents that eliminates the need for manually collected and labeled offline training data. By leveraging vision-language models (VLMs), \name{} enables automatic generation of training tasks and reward signals within GUI environments, removing human cost from both task design and evaluation.
We further introduce a two-stage reinforcement learning strategy: training on generated tasks to acquire general capabilities, followed by test-time adaptation using VLM-based rewards.
We conduct extensive experiments across different base models (Aguvis-7B and UI-TARS-7B-DPO) and different GUI environments. Notably, \name{}-Aguvis-7B achieves a 63\% relative improvement, while \name{}-UI-TARS-7B shows a 14\% relative improvement.

\textbf{Acknowledgements. }The work is supported by the National Key R\&D Program of China (NO. 2022ZD0161300), by the National Natural Science Foundation of China (U24A20325, 62321005, 62376134).

\appendix

\section{Implementation Details}

\subsection{Task Generation}

We use GPT-4o~\cite{hurst2024gpt} for task generation.
Each prompt consists of task exemplars and the screenshot of a randomly sampled initial state.
To encourage diversity, each prompt generates multiple task candidates: 10 per prompt in OSWorld~\cite{xie2024osworld} and 5 per prompt in AndroidLab~\cite{xu2024androidlab}.
In total, we generate over 4,000 Ubuntu-based tasks and 225 Android-based tasks, from which we randomly sample 725 and 175 tasks, respectively, as the final training set.
The prompt template used to generate OSWorld tasks is shown below:

\begin{tcolorbox}[title=Task Generation Prompt Template, listing only, listing options={style=promptstyle}, breakable, before upper=\ttfamily\small, boxrule=0.3pt, top=5pt, bottom=5pt, left=4pt, right=4pt,]

You will be given a screenshot of the current state of the computer. 

Based on the screenshot, generate \{Number of Instructions\} human-like instructions for an Ubuntu-based task related to using \{App Name\} app. 

1. The task should be something that **can be achieved** using common interactions on a standard Ubuntu system from the current state.

2. The proposed task should be similar to the example tasks but not exactly the same.

3. The \{Number of Instructions\} instructions should cover different aspects of using \{App Name\}. During the generation, think about any other kind of tasks differ from the examples you have generated before.

4. The tasks should be clear and unambiguous, with specific and detailed instructions, and can be evaluated objectively by observing the action trajectory. 

5. If the task you propose requires access to any internet resources, you must ensure that the URL or resource exists.

6. The task you propose must require at least 3 steps to complete, and the difficulty and the complexity should be evenly distributed.
\newline

The following are examples of good task instructions (their initial screenshots/states are different):

\{List of Exemplars\}
\newline

Please follow the **content, type, and distribution** of the examples provided to you, and generate task instructions based on the initial screenshot.

**Do not repeat the existing examples.**
\newline

Now, you will be given a initial computer state screenshot. Please generate \{Number of Instructions\} task instructions based on the screenshot.

Return only the task instruction text. Each instruction should occupy a single line. \{Number of Instructions\} lines of instructions should be returned.
\end{tcolorbox}

\subsection{Reward Estimation}

For efficient reward estimation, we use a locally deployed QwenVL-2.5-32B-Instruct~\cite{Qwen2.5-VL}.
The VLM assigns binary rewards (success/failure) to the agent's trajectories.
The VLM is prompted with all screenshots of the trajectory and the task instruction. 
To improve reliability, we use a voting mechanism: the VLM is queried multiple times with temperature 1.0, and the final reward is based on unanimous agreement.
The reward estimation prompt for OSWorld tasks is shown below:

\begin{tcolorbox}[title=Reward Estimation Prompt Template, listing only, listing options={style=promptstyle}, breakable, before upper=\ttfamily\small, boxrule=0.3pt, top=5pt, bottom=5pt, left=4pt, right=4pt,]

You will be given a task instruction and a series of screenshots of the task execution. 

Please analyze the screenshots and provide a detailed analysis of the task completion by following the steps below:

1. First, analyze and understand the task instruction. Describe what should the screenshots look like if the task is completed successfully.

2. Describe what you observe in each screenshot, analysis what actions were taken and what changes were made to the UI to achieve the task (or mistakes made).

3. When you analyze the screenshots, please pay attention to the very detailed elements and changes in the UI. Every small detail may affect the final result.

4. After all screenshots are analyzed, provide a overall reasoning about how the task was completed or failed at **the final state**. Make sure you have considered all demands of the task instruction.

5. Determine if the task was completed at **the final state** (the last screenshot) successfully (score 1 for success, 0 for failure). If the task is completed during the process but not at the final state, it should be considered as failure (0 score).
\newline

Provide your response strictly in the following format:

TASK REQUIREMENT:

[Your understanding of the task instruction]
\newline

SCREENSHOT ANALYSIS:

Screenshot 1:

[Analysis of first screenshot]

Screenshot 2:

[Analysis of second screenshot]

...
\newline

REASONING:

[Your reasoning]
\newline

FINAL ANSWER:

[Your final answer]
\newline

SCORE: [0/1]
\newline

Here is an example:

(Task Instruction: Please help me backup my emails in "Bills" folder in Thunderbird and store the .eml files with only subject names to my Google Drive folder called "emails".)
\newline

TASK REQUIREMENT:

- Backup the emails in "Bills" folder in Thunderbird.

- Store the backup .eml files with only subject names, and the emails should be saved in the Google Drive folder called "emails".

- Once succeed, the emails should be visible in the Google Drive folder "emails". Or at least there should be a saving action performed.
\newline

SCREENSHOT ANALYSIS:
\newline

Screenshot 1:

- Thunderbird email client is open.

- The "Bills" folder is visible under "Local Folders."

- There is no observable action performed yet in this screenshot.
\newline

Screenshot 2:

- The "Bills" folder has been selected, and the folder content is displayed.

- Two emails are visible: "Amazon Web Services Invoice Available" and "Your receipt from X (formerly Twitter)."

- No further actions are taken on the emails.
\newline

Screenshot 3:

- Both emails in the "Bills" folder are selected.

- Content previews of both emails are displayed on the right-hand side.

- No observable attempt to export or save the emails is visible.
\newline

Screenshot 4:

- The right-click context menu is accessed for the selected emails.

- The "Save As..." option is hovered over, indicating intent to save the selected emails.
\newline

Screenshot 5:

- The file navigation window opens, allowing the user to choose a save destination.

- No specific Google Drive folder (e.g., "emails") is accessed or visible in this screenshot.
\newline

Screenshot 6:

- The "Desktop" option in the file picker is hovered over.

- Still no indication of Google Drive folder ("emails") selection.
\newline

Screenshot 7:

- The "Show other locations" option is hovered over in the file picker.

- No confirmation that the user is navigating to Google Drive or saving the files with subject names only.
\newline

Screenshot 8:

- The "Software Updates Available" notification appears. The file picker is still open without any observable confirmation of file saving or destination selection.

- It remains unclear where or if the emails have been saved.
\newline

REASONING:

Based on the screenshots provided:

1. While there was some intent to save the emails (as shown by the selection and access of the "Save As..." function), there is no confirmation that the .eml files were saved with subject names only and placed in the required Google Drive folder ("emails").

2. The screenshots lack evidence of the completion of the task as per the instructions.
\newline

FINAL ANSWER:

The task was not completed successfully due to the lack of observable saving action.
\newline

SCORE: 0
\newline

Now, please **strictly follow the format** and analyze the following screenshots (The last line should only be SCORE: [0/1], no other text):

Task Instruction: \{instruction\}

Screenshots (by order): \{screenshots\}
\end{tcolorbox}

\subsection{RL Training}
\setlength{\tabcolsep}{2pt}
\setlength{\doublerulesep}{2\arrayrulewidth}
\renewcommand{\arraystretch}{1.1}

\begin{table}[h]
    \renewcommand{\thetable}{5}
    \centering
    \caption{\textbf{Training hyper-parameters of \name{}.}}
    \label{table:hyper}
    
    \begin{tabular}{lc}
        \toprule
        GRPO group size & 64 \\
        KL loss coefficient & 0.1 \\
        rollout temperature & 0.5 \\
        rollout top-p & 0.9 \\
        rollout frequency penalty & 1.0 \\
        train batch size & 16384 \\
        optimizer & AdamW \\
        learning rate & 2e-6 \\
        lr schedule & constant \\
        weight decay & 0.0 \\
        optimizer momentum & $\beta_1, \beta_2=0.9, 0.95$ \\
        \bottomrule
    \end{tabular}
\end{table}

Hyper-parameters of our RL training are listed in Tab.~\ref{table:hyper}.

\section{Analysis of KL Loss}
\subsection{Theoretical Derivation and Analysis}
Here, we derive the gradient from the KL loss and show why our modification benefits stable training.

In our implementation, the model has only a single update after each
exploration stage, ensuring that $\pi_\theta=\pi_{\theta_{\text{old}}}$.
Therefore, Eq.~\ref{eq:loss} can be simplified by removing the min and clip operation:

\begin{equation}
\mathcal{J}(\theta)=\frac{1}{|a|}\sum_{k=1}^{|a|}\left(\frac{\pi_\theta(a_k)}{\pi_{\theta_{\text{old}}}(a_k)}\hat{A} -\beta D_{\text{KL}}(\pi_\theta||\pi_{\text{ref}})\right)
\tag{8}
\end{equation}

where $a$ denotes a predicted action sequence with the corresponding advantage $\hat{A}$, $\pi_\theta(a_k)$ stands for $\pi_\theta\left(a_k|I, s, a_{<k}\right)$, $I$ is the task instruction and $s$ is the agent's state.

\noindent\textbf{GRPO.} GRPO~\cite{shao2024deepseekmath} uses a per-token k3-estimator~\cite{schulman2020kl} for the KL loss:

\begin{equation}
\label{eq:kl_grpo_appendix}
D_{\text{KL}}^{\text{GRPO}}(\pi_\theta||\pi_{\text{ref}})=\frac{\pi_{\text{ref}}(a_k)}{\pi_\theta(a_k)}-\log \frac{\pi_{\text{ref}}(a_k)}{\pi_\theta(a_k)} -1
\tag{9}
\end{equation}

The gradient with respect to the parameter $\theta$ is:

\begin{equation}
\nabla_\theta\mathcal{J}^{\text{GRPO}}(\theta)=\frac{1}{|a|}\sum_{k=1}^{|a|}\left(\hat{A}+\beta\left(\frac{\pi_{\text{ref}}(a_k)}{\pi_\theta(a_k)}-1\right)\right)\nabla_\theta\log\pi_\theta(a_k)
\tag{10}
\end{equation}

Referring to GRPO, the Gradient Coefficient of the KL loss can be written as:

\begin{equation}
GC_{\text{KL}}^{\text{GRPO}}(a_k)=\frac{\pi_{\text{ref}}(a_k)}{\pi_\theta(a_k)}-1
\tag{11}
\end{equation}

\noindent\textbf{Ours.} We replace the original KL loss in GRPO with the k2-estimator~\cite{schulman2020kl}, \ie, per-token MSE loss:

\begin{equation}
\label{eq:kl_appendix}
D_{\text{KL}}(\pi_\theta||\pi_{\text{ref}})=\frac{1}{2}\left(\log\pi_\theta(a_k)-\log\pi_{\text{ref}}(a_k)\right)^2
\tag{12}
\end{equation}

The gradient becomes:
\vspace{-0.5em}

\begin{equation}
\nabla_\theta\mathcal{J}(\theta)=\frac{1}{|a|}\sum_{k=1}^{|a|}\left(\hat{A}+\beta\left(\log\pi_{\text{ref}}(a_k)-\log\pi_\theta(a_k)\right)\right)\nabla_\theta\log\pi_\theta(a_k)
\tag{13}
\end{equation}

\begin{equation}
GC_{\text{KL}}(a_k)=\log\pi_{\text{ref}}(a_k)-\log\pi_\theta(a_k)
\tag{14}
\end{equation}

\begin{figure}[t]
\renewcommand{\thefigure}{4}
    \centering
    \includegraphics[width=0.48\linewidth]{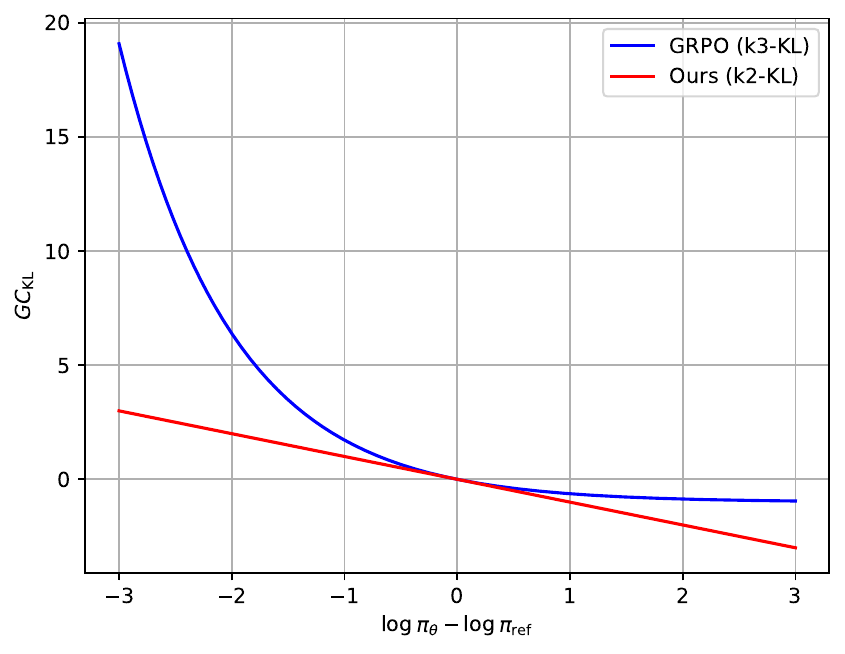}
    \vspace{-0.1em}
    \caption{Gradient coefficient of KL loss.}
    \label{fig:kl_gc}
    \vspace{-0.5em}
\end{figure}

\begin{figure}[t]
\renewcommand{\thefigure}{5}
    \centering
    \begin{subfigure}[b]{0.32\linewidth}
        \includegraphics[width=\linewidth]{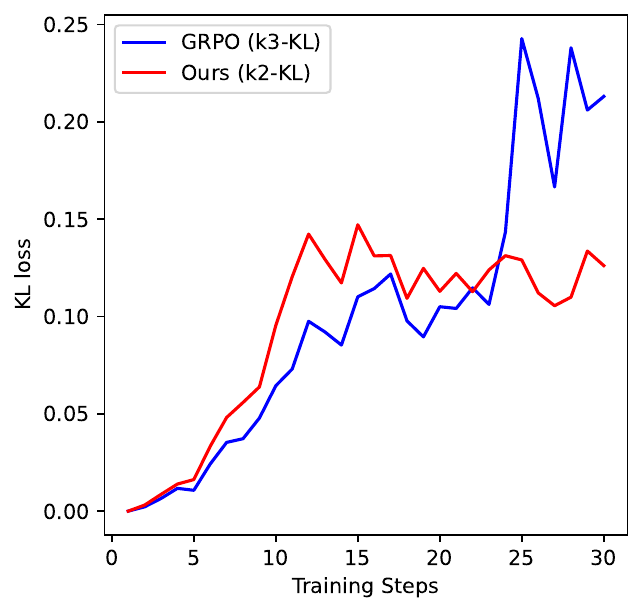}
        \caption{KL loss.}
    \end{subfigure}
    \hfill
    \begin{subfigure}[b]{0.31\linewidth}
        \includegraphics[width=\linewidth]{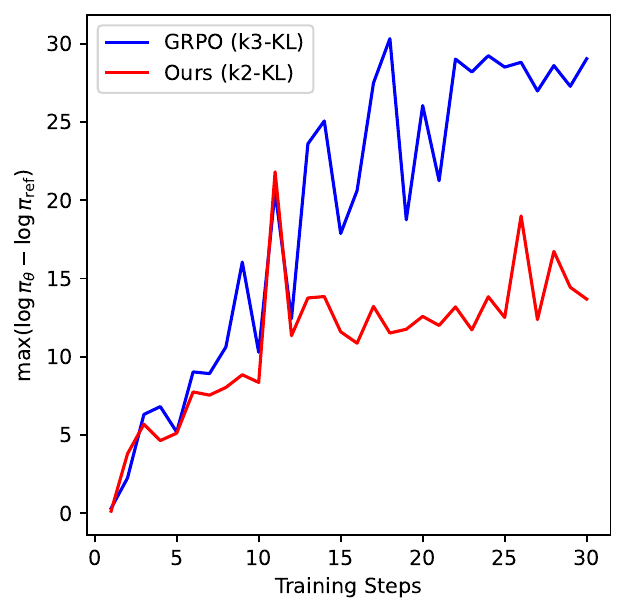} 
        \caption{$\max(\log\pi_\theta-\log\pi_{\text{ref}})$}
    \end{subfigure}
    \hfill
    \begin{subfigure}[b]{0.32\linewidth}
        \includegraphics[width=\linewidth]{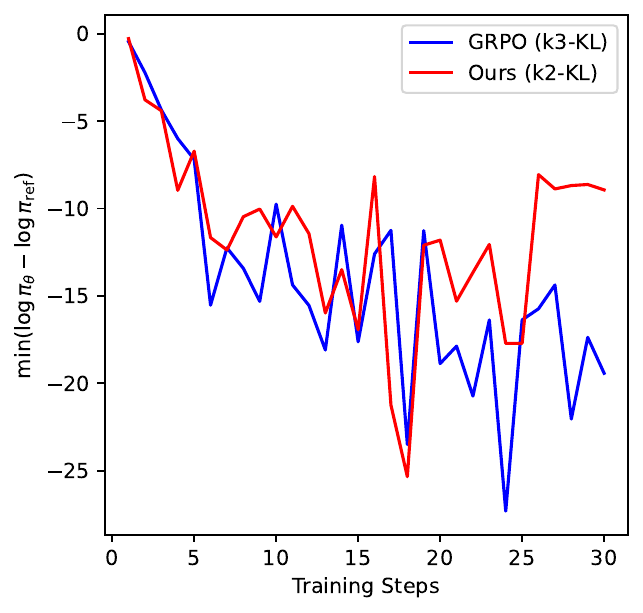}
        \caption{$\min(\log\pi_\theta-\log\pi_{\text{ref}})$}
        \label{fig:kl_min_curve}
    \end{subfigure}
    \caption{KL loss curve and token-wise maximum and minimum of $\log\pi_\theta-\log\pi_{\text{ref}}$ during training.}
    \label{fig:kl_curve}
    \vspace{-1em}
\end{figure}

\noindent\textbf{Analysis.} We plot the curve of the gradient coefficients $GC_{\text{KL}}$ and $GC_{\text{KL}}^{\text{GRPO}}$ in Fig.~\ref{fig:kl_gc}, with $\log\pi_\theta-\log\pi_{\text{ref}}$ as the x-axis.
We also record the KL loss and the token-wise maximum and minimum of $\log\pi_\theta-\log\pi_{\text{ref}}$ during training, as shown in Fig.~\ref{fig:kl_curve}.
As the training model $\theta$ gradually deviates from the reference model, there exist some tokens such that $\log\pi_\theta\ll\log\pi_{\text{ref}}$ (Fig.~\ref{fig:kl_min_curve}).
For GRPO, this leads to a huge gradient coefficient, which may cause training instability.
In contrast, ours has a stable gradient coefficient and avoids large gradients.
Some concurrent works~\cite{Liu2025KL} have also reached derivations and conclusions similar to ours.

Furthermore, in practical implementations, language models typically output log probabilities, \ie, $\log \pi$, computed from logits.
When computing the KL loss of GRPO (Eq.~\ref{eq:kl_grpo_appendix}), exponentiation is required, which is prone to overflow or underflow.
In contrast, our KL loss (Eq.~\ref{eq:kl_appendix}) avoids this issue.

\subsection{Supplementary Ablation Study}
\vspace{-0.6em}

\begin{figure}[ht]
\renewcommand{\thefigure}{6}
    \centering
    \includegraphics[width=0.4\linewidth]{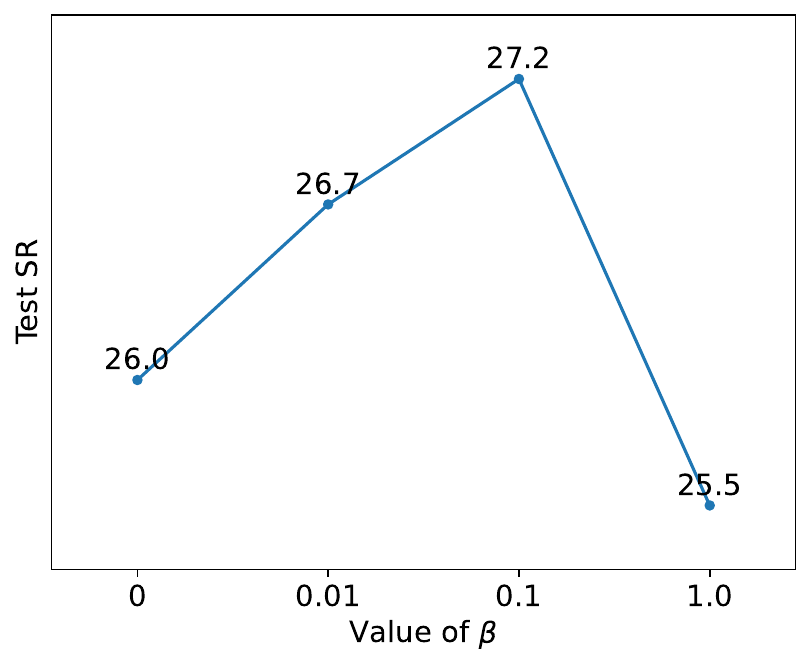}
    \vspace{-0.2em}
    \caption{Test success rates with different KL loss coefficients $\beta$.}\
    \label{fig:kl_beta}
    \vspace{-0.8em}
\end{figure}

We compare different KL loss coefficients $\beta$, and the results shown in Fig.~\ref{fig:kl_beta}.
Although some existing work~\cite{yu2025dapo,liu2025understanding} suggests removing the KL penalty for general reasoning tasks, our findings differ in the context of training GUI agents.
We observe that setting $\beta$ yields the best test performance.
Removing the KL loss entirely ($\beta=0$) or using a small $\beta$ (\eg, 0.01) leads to performance degradation, likely due to the policy distribution drift that the model overfits to current tasks.
In contrast, a large $\beta$ (\eg, 1) imposes excessive constraints on optimization, also resulting in worse results.

\section{Analysis of Task Generation}
\subsection{Task Example}
\setlength{\tabcolsep}{2pt}
\setlength{\doublerulesep}{2\arrayrulewidth}
\renewcommand{\arraystretch}{1.1}

\newcolumntype{M}[1]{>{\centering\arraybackslash}m{#1}}
\newcolumntype{L}[1]{>{\raggedright\arraybackslash}m{#1}}

\begin{table}[ht]
\renewcommand{\thetable}{6}
\centering
\small
\caption{Examples of generated tasks and test tasks on OSWorld}
\label{tab:osworld_task}

\renewcommand{\arraystretch}{1.4}
\begin{tabular}{M{1.9cm} L{5.8cm} L{5.8cm}}
\toprule
\textbf{Domain} & \textbf{Generated Task Instructions} & \textbf{Test Task Instructions} \\

\midrule
Chrome
& Look up ``MIT's Deep Learning State of the Art lecture'' mentioned in the article and play the video if it's available.
&  Find a men's T-Shirt that is in large size with a stripe pattern, short sleeve and under the Sales\&Discount.
\\

\midrule
GIMP
& Could you add a new layer to the image, call it ``Overlay'', and set its mode to ``Multiply''?
& Could you make the background of this image transparent for me?
\\

\midrule
Office
& Could you create a new column called ``Average Sales'' for each product in Zones 1, 2, and 3 by calculating the average of Q1, Q2, Q3, and Q4?
\vspace{0.3em}\newline 
Could you add the hyperlink ``https://its.example.com'' to the text ``Information Technology Services (ITS)'' in the main content?
& I would like to pad all the numbers in the `Old ID' column with zeros in front, to fill them up to seven digits in the `New 7 Digit ID' column.
\vspace{0.3em}\newline 
Export the current document into PDF, keep the file name
\\

\midrule
OS
& Could you create a folder called ``WorkProjects'' in your home directory and put a copy of ``installed\_packages.txt'' inside it?
& Use terminal command to count all the lines of all php files in current directory recursively, show the result on the terminal
\\

\midrule
VLC
& Could you adjust the playback speed in VLC to 1.5x for this video? I want to finish it faster, even if the audio isn't perfect.
& Help me modify the folder used to store my recordings to Desktop
\\

\midrule
VS Code
& I'd like to stop the Welcome Page from showing up every time I open VS Code. Can you help me with that?
& Please help me use VS Code to open the ``project'' in the ``user'' folder under ``home''.
\\
\bottomrule
\end{tabular}
\end{table}
\setlength{\tabcolsep}{2pt}
\setlength{\doublerulesep}{2\arrayrulewidth}
\renewcommand{\arraystretch}{1.1}

\newcolumntype{M}[1]{>{\centering\arraybackslash}m{#1}}
\newcolumntype{L}[1]{>{\raggedright\arraybackslash}m{#1}}

\begin{table}[ht]
\renewcommand{\thetable}{7}
\centering
\small
\caption{Examples of generated tasks and test tasks on AndroidLab}
\label{tab:android_task}

\renewcommand{\arraystretch}{1.4}
\begin{tabular}{M{1.9cm} L{5.8cm} L{5.8cm}}
\toprule
\textbf{Apps} & \textbf{Generated Task Instructions} & \textbf{Test Task Instructions} \\

\midrule
Clock
& Set an alarm for 5:30AM every Tuesday and Thursday, label it as ``Workout'', and disable vibrate.
& I need to set a 10:30PM clock every weekend, and label it as ``Watch Football Games'' to remind me.
\vspace{0.3em}\newline 
Does my alarm at 4PM turn on vibrate? \\

\midrule
Calendar
& Arrange an event titled ``project deadline'' on March 15th, and attach a note saying ``Submit by noon.''
& I want to add an event at 5:00PM today, whose Title is ``work''. \\

\midrule
Bluecoins
& Modify the transaction on May 3, 2024, from ``expense'' to ``income'', set the amount to 780 CNY, and update the note to ``Refund''.
& Change the type of the transaction on May 2, 2024, from ``income'' to``expense'', adjust the amount to 520 CNY, and change the note to ``Wrong Operation''. \\

\midrule
Contacts
& Set Xu's birthday to be 1989/06/15 and add his website as xuportfolio.com.
& Add birthday to AAA as 1996/10/24
\vspace{0.3em}\newline 
Set contacts ABC's website to be abc.github.com \\

\midrule
Setting
& Set screen timeout duration to 5 minutes.
\vspace{0.3em}\newline 
Enable the auto-rotate screen feature.
& Turn on airplane mode of my phone
\vspace{0.3em}\newline 
Turn my phone to Dark theme \\

\bottomrule
\end{tabular}
\end{table}

As shown in Tab.~\ref{tab:osworld_task} and Tab.~\ref{tab:android_task}, we list some of the generated tasks and the test tasks on OSWorld and AndroidLab for comparison. 

The generated tasks leverage the world knowledge embedded in VLMs to extend common operations, thereby enhancing data diversity. For example, the model generates Android tasks such as enabling auto-rotate and setting screen timeout, and Ubuntu tasks like speeding up video playback—features not present in the original test tasks.

Additionally, the generated tasks are both feasible and well-grounded. For instance, tasks like copying a package list file to a new project folder and averaging sales data for Ubuntu, or setting reasonable alarms and editing calendars as humans would for Android, ensure practical applicability.

\subsection{Visualization}
\begin{figure}[ht]
    \renewcommand{\thefigure}{7}
    \centering
    \includegraphics[width=0.6\linewidth]{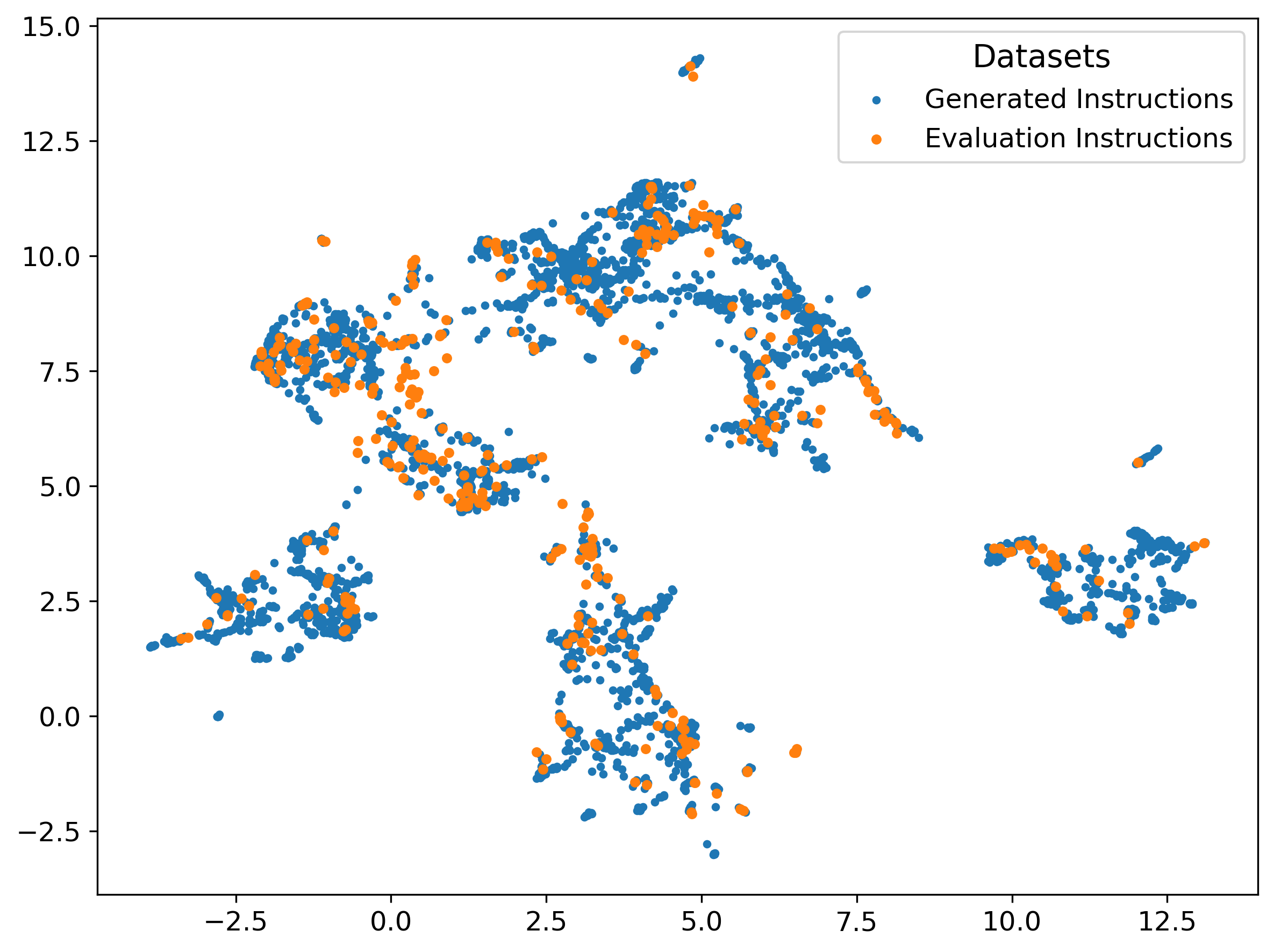}
    \caption{Visualization of the test and generated task instructions in OSWorld.}
    \label{fig:ins_embed}
\end{figure}

We visualize the task instructions in the original test set and our generated set.
Following OS-Genesis~\cite{sun2024genesis}, we use Sentence-BERT~\cite{reimers2019sentence} to encode instructions into a high-dimensional semantic space, which is then reduced to two dimensions using UMAP~\cite{mcinnes2018umap}.
As shown in Fig.~\ref{fig:ins_embed}, orange points represent instructions from the original OSWorld test set, while blue points correspond to instructions generated by our method.

Each cluster in the visualization corresponds to a distinct application domain (e.g., specific applications or task types), reflecting strong semantic coherence within domains. 
Notably, the generated instructions, conditioned on exemplars, not only align well with their target domains, but also extend into previously unexplored regions of the instruction space.
This illustrates the capacity of our method to produce semantically valid and diverse tasks beyond the original dataset.

\section{Qualitative Comparisons}
We conduct case studies to further demonstrate the effectiveness of our proposed \name{}.
We observe that the base model UI-TARS-7B-DPO~\cite{qin2025ui} shows limited task comprehension and inadequate attention to detail, frequently falling into repetitive action loops during task execution. In contrast, after our \name{} training, the model exhibits significantly more stable behavioral strategies and stronger task execution capabilities.

\begin{figure}[ht]
\centering
\renewcommand{\thefigure}{8}
  \includegraphics[width=\textwidth]{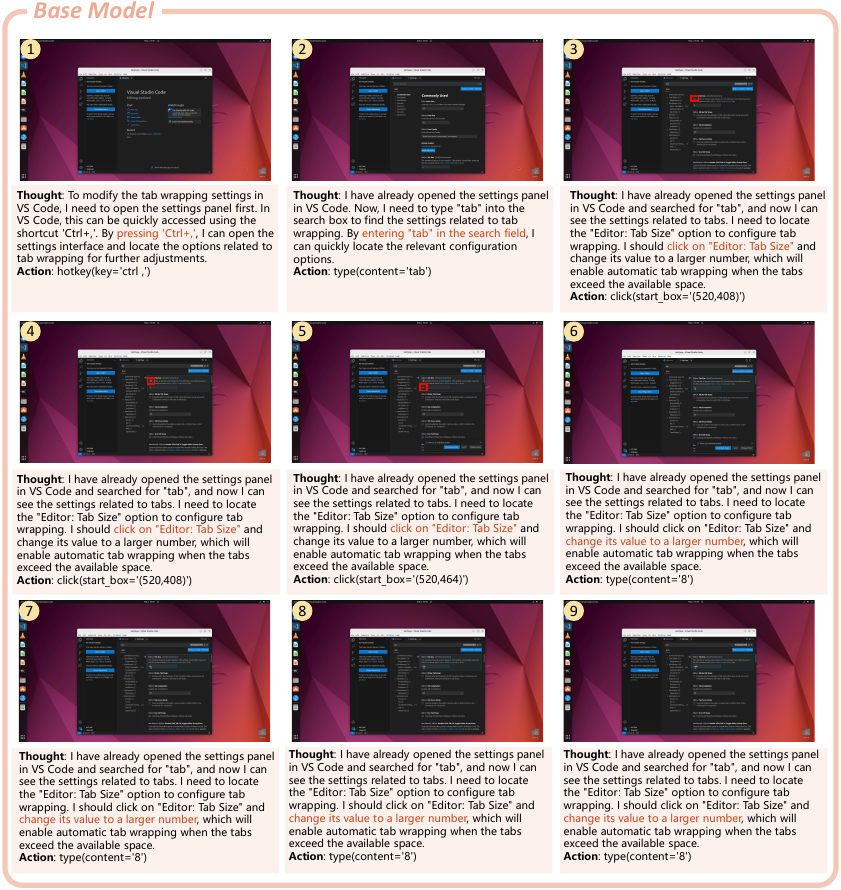}
  \captionof{figure}{\textbf{The trajectory of the base model UI-TARS-7B-DPO when executing a VS Code task.} The instruction is \emph{``I want the tabs to wrap onto multiple lines when they exceed the available space. Please help modify the VS Code settings accordingly.''} The original thoughts are in Chinese and have been translated into English for presentation purposes. Subsequent repetitive steps have been omitted.}
  \label{fig:vscode-1}
\end{figure}

\begin{figure}[ht]
\centering
\renewcommand{\thefigure}{9}
  \includegraphics[width=\textwidth]{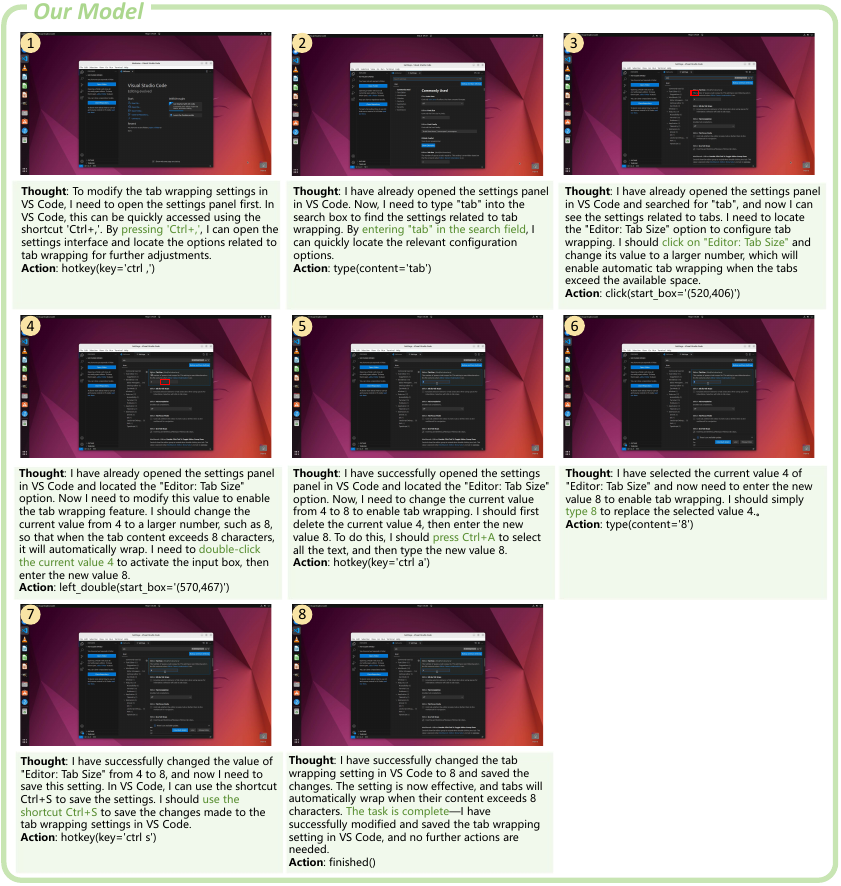}
  \captionof{figure}{\textbf{The trajectory of our trained model when executing a VS Code task.} The instruction is \emph{``I want the tabs to wrap onto multiple lines when they exceed the available space. Please help modify the VS Code settings accordingly.''} The original thoughts are in Chinese and have been translated into English for presentation purposes.}
  \label{fig:vscode-2}
\end{figure}

For example, as shown in Fig.~\ref{fig:vscode-1} and~\ref{fig:vscode-2}, a task in the VS Code domain of OSWorld involves the instruction: \emph{``I want to make the tabs wrapped over multiple lines when exceeding available space, please help modify the setting of VS Code.''} During execution, the base model attempts to modify the ``Tab Size'' parameter but fails to delete the default value before entering a new one. Instead, it prepends the new number to the existing value, resulting in an incorrect setting. This faulty operation is then repeated multiple times, indicating that the model lacks the ability to detect invalid actions. In contrast, our model adopts a more robust action strategy: it first uses a keyboard shortcut to select all existing content, then inputs the correct value, and successfully completes the task.

\begin{figure}[ht]
\centering
\renewcommand{\thefigure}{10}
  \includegraphics[width=\textwidth]{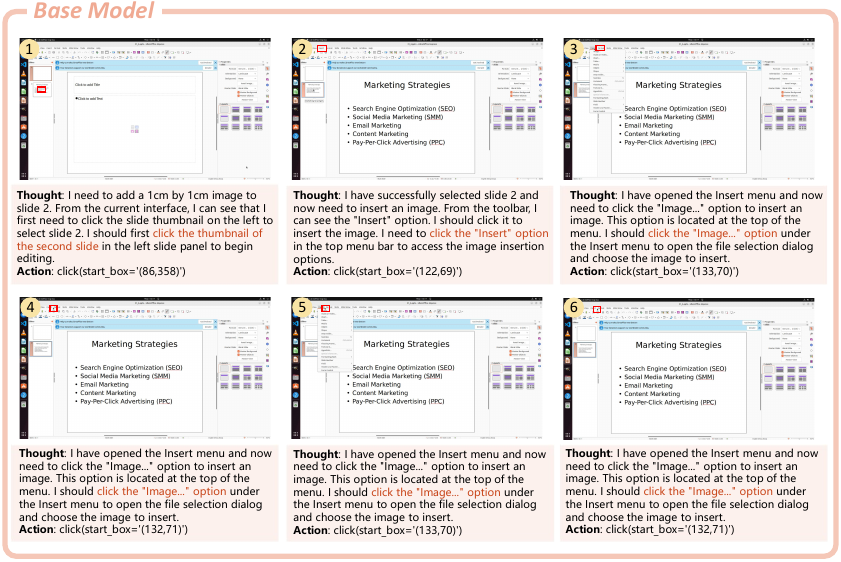}
  \captionof{figure}{\textbf{The trajectory of the base model UI-TARS-7B-DPO when executing a LibreOffice Impress task.} The instruction is \emph{``Add an image `none.png' on the Desktop to slide 2 with 1cm*1cm size.''} The original thoughts are in Chinese and have been translated into English for presentation purposes. Subsequent repetitive steps have been omitted.}
  \label{fig:impress-1}
\end{figure}

\begin{figure}[ht]
  \centering
  \renewcommand{\thefigure}{11}
  \includegraphics[width=\textwidth]{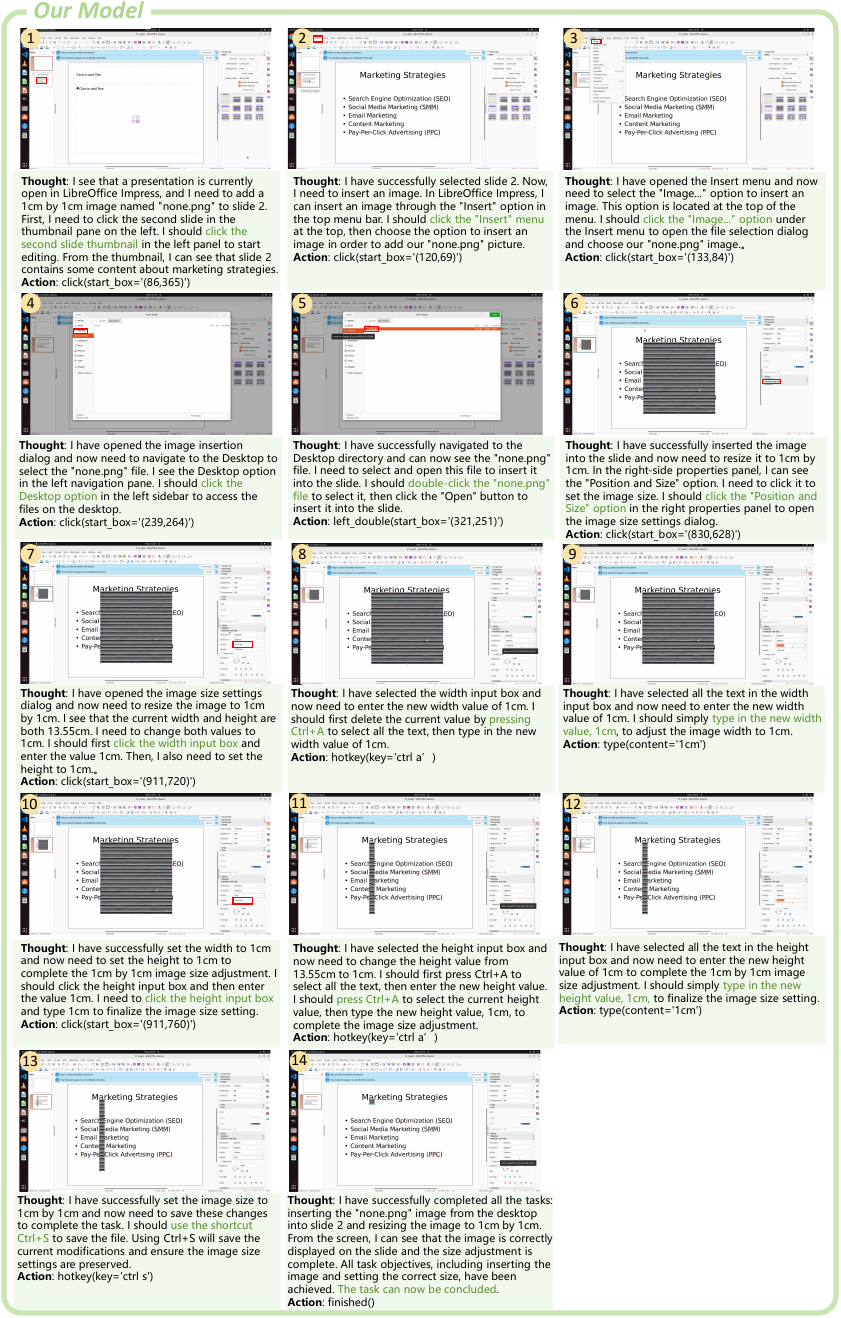}
  \caption{\textbf{The trajectory of our trained model when executing a LibreOffice Impress task.} The instruction is \emph{``Add an image `none.png' on the Desktop to slide 2 with 1cm*1cm size.''} The original thoughts are in Chinese and have been translated into English for presentation purposes.}
  \label{fig:impress-2}
\end{figure}

Another example comes from a task in the LibreOffice Impress domain, where the instruction is: \emph{``Add an image `none.png' on the Desktop to slide 2 with 1cm*1cm size.''} As shown in Fig.~\ref{fig:impress-1} and~\ref{fig:impress-2}, after clicking the ``Insert'' menu, the base model attempts to select the ``Image'' option but misclicks a blank area due to inaccurate grounding, causing the menu to close prematurely. However, the model fails to detect this change and continues to attempt to click the ``Image'' option under the now-closed ``Insert'' menu, resulting in ineffective repetition. In contrast, our model completes the full insertion process more reliably. It successfully opens the image insertion interface, selects the correct image file, adjusts the width and height step by step, and ultimately finishes the long-horizon task accurately. This comparison further demonstrates the improved stability and general performance of our method in handling complex tasks.

\begin{figure}[ht]
\centering
\renewcommand{\thefigure}{12}
  \includegraphics[width=\textwidth]{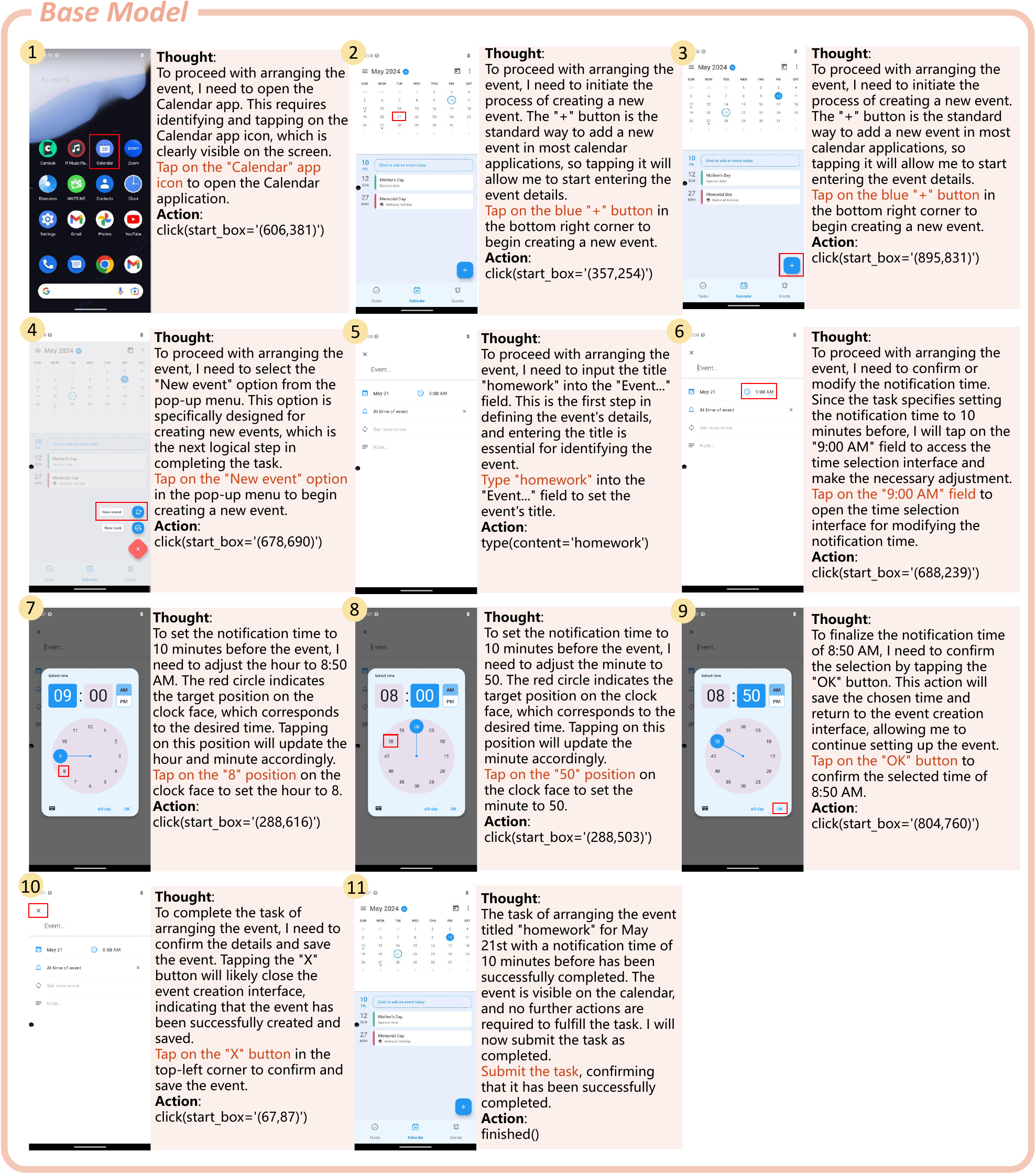}
  \captionof{figure}{\textbf{The trajectory of the base model UI-TARS-7B-DPO when executing a Calendar task.} The instruction is \emph{``You should use calendar to complete the following task: Arrange an event titled `homework' for me at May 21st, and set the notification time to be 10 minutes before.''}}
  \label{fig:calendar-1}
\end{figure}

\begin{figure}[ht]
\centering
\renewcommand{\thefigure}{13}
  \includegraphics[width=\textwidth]{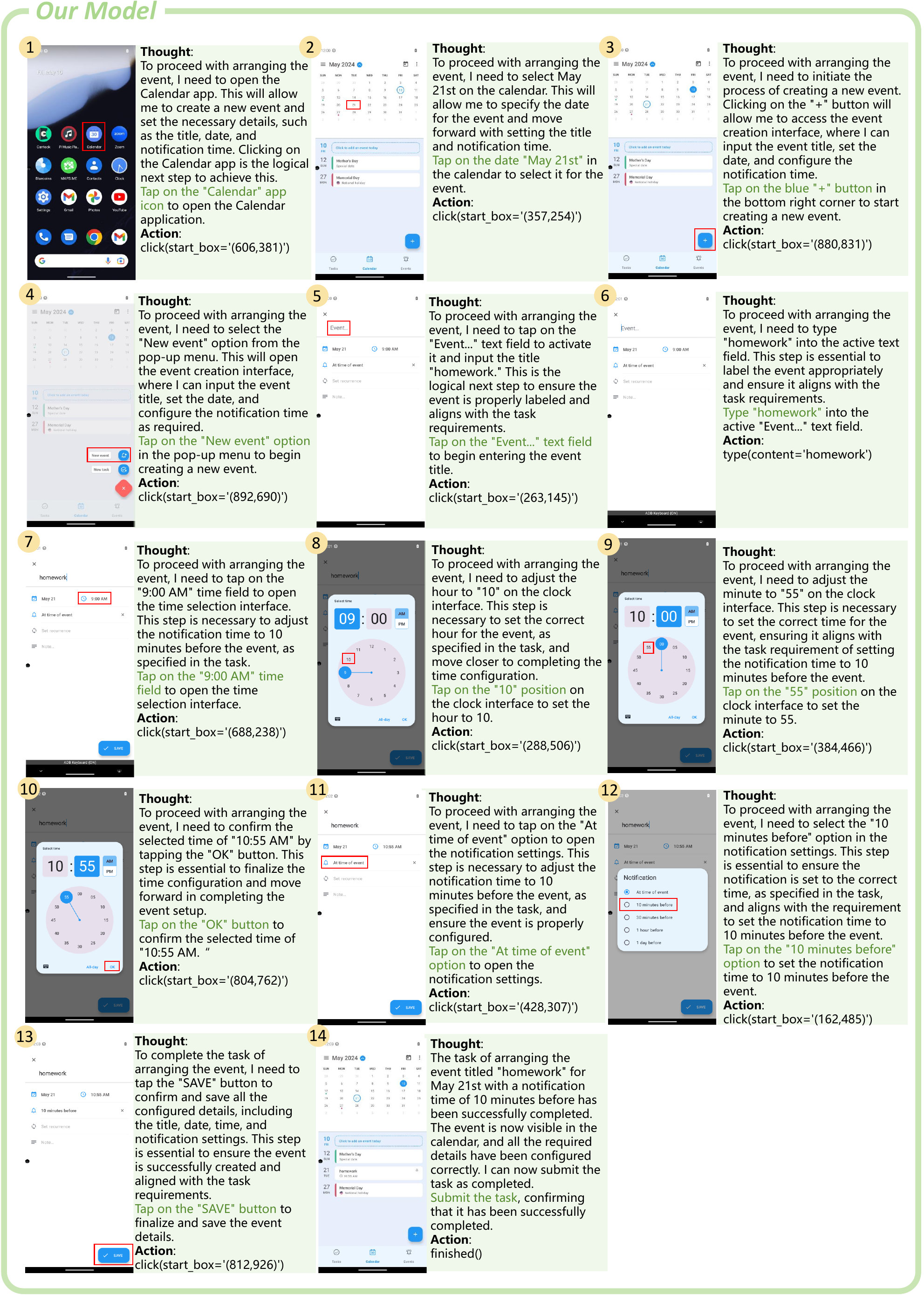}
  \captionof{figure}{\textbf{The trajectory of our trained model when executing a Calendar task.} The instruction is \emph{``You should use calendar to complete the following task: Arrange an event titled `homework' for me at May 21st, and set the notification time to be 10 minutes before.''}}
  \label{fig:calendar-2}
\end{figure}

When executing tasks in AndroidLab, the model also shows significant performance improvements. Take a Calendar task as an example, with the instruction: \emph{``You should use calendar to complete the following task: Arrange an event titled `homework' for me at May 21st, and set the notification time to be 10 minutes before.''}, as illustrated in Fig.~\ref{fig:calendar-1} and~\ref{fig:calendar-2}. The base model shows insufficient understanding of the instruction, overlooking details such as the event title and notification time. It neither adds the event title nor sets the notification, and closes the interface ultimately. In contrast, our trained model accurately captures and executes these detailed requirements, successfully adding the event on the specified date and setting the notification, which demonstrates the effectiveness of our training framework in mitigating issues arising from overlooked details and improving task accuracy.

It is worth noting that the base model fails to complete any of the above tasks, indicating the lack of supervision from successful trajectories. Through training on generated tasks, our model learns more generalizable interaction strategies. This capability not only enhances its performance on specific tasks, but also demonstrates the significant potential and practical effectiveness of the proposed training framework in enabling the model to adapt to complex GUI environments.

\clearpage

\null
\vfill
\vfill
\null

\clearpage

\null
\vfill
\vfill
\null

\clearpage

\null
\vfill
\vfill
\null

\clearpage

\clearpage

\null
\vfill
\vfill
\null

\clearpage

\clearpage
\bibliographystyle{abbrvnat}
{\small
\bibliography{egbib}
}

\end{document}